\documentclass{article}
\usepackage[preprint]{neurips_2026}

\usepackage[utf8]{inputenc}
\usepackage[T1]{fontenc}
\usepackage[colorlinks=true,linkcolor=black,citecolor=black,urlcolor=blue,breaklinks=true]{hyperref}
\usepackage{url}
\usepackage{booktabs}
\usepackage{amsfonts}
\usepackage{amsmath}
\usepackage{amssymb}
\usepackage{nicefrac}
\usepackage{microtype}
\usepackage{xcolor}
\usepackage{graphicx}
\usepackage{rotating}
\usepackage{todonotes}

\title{Is Capability a Liability? More Capable Language Models Make Worse Forecasts When It Matters Most}

\author{%
  Nick Merrill \\
  Forecasting Research Institute \\
  UC Berkeley \\
  \texttt{nick@forecastingresearch.org} \\
  \And
  Jaeho Lee \\
  Forecasting Research Institute \\
  \texttt{jaeho@forecastingresearch.org} \\
  \And
  Ezra Karger \\
  Forecasting Research Institute \\
  \texttt{ezra@forecastingresearch.org} \\
}

\begin{document}

\maketitle

\begin{abstract}
We document inverse scaling in LLMs on forecasting problems whose underlying time series exhibit superlinear growth and tail risk of regime change, a structure common in finance and epidemiology. On these tasks, more capable models produce worse distributional forecasts. The pattern appears on ForecastBench-Sim (FBSim), a contamination-free, simulated-world benchmark we release, in forecasting synthetic SIR epidemics with a matched linear control, and replicates in real-world datasets on COVID-19, measles, housing markets, and hyperinflation. A per-quantile decomposition shows the failure concentrates at the upper tail, which more capable models shift upward to track aggressive extrapolations of growth, while the lower tail stays put. A within-family study of Llama-3.1 shows that both model scale and post-training independently contribute to this effect. Domain knowledge does not reliably rescue calibration. This inverse scaling does not appear on single-threshold metrics common in LLM forecasting benchmarks, reversing the sign of the capability--accuracy relationship on identical outputs. Single-threshold scoring at conventional cutoffs misses the upper-tail cost; tail-inclusive scoring reverses the sign of the capability--accuracy relationship on the same outputs. We recommend that LLM forecasting evaluations use continuous (and unbounded) measures of accuracy alongside bounded binary threshold metrics.
\end{abstract}

\begin{figure}[!h]
\centering
\includegraphics[width=\linewidth]{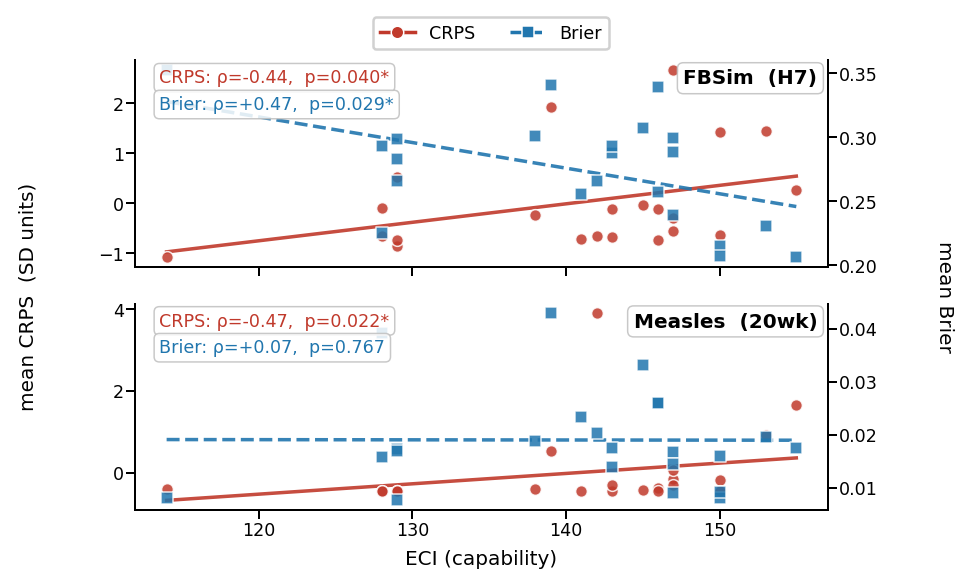}
\caption{On a continuous proper scoring rule (red), more capable models produce worse forecasts in two, unrelated domains. This inverse scaling does not appear on a threshold-based proper scoring rule (blue). \textbf{Top}: FBSim, pooled. \textbf{Bottom}: pre-vaccine measles. Both series shown at furthest forecast horizons.}
\label{fig:hero}
\end{figure}

\section{Introduction}
\label{sec:intro}

 Larger language models (LLMs) tend to perform better across almost all tasks \citep{kaplan2020scaling, chowdhery2023palm}, implying that increasing scale will bring general improvements. For this reason, a growing body of work studies the exceptions: inverse scaling, cases in which more capable models perform worse \citep{mckenzie2023inverse, wei2023ushaped, zhou2024larger}. We demonstrate a new, structurally identifiable class of inverse scaling: distributional forecasting on time series with superlinear growth and tail risk of regime change. Unlike previously documented inverse-scaling tasks, which are typically adversarial, narrow, or dependent on a particular prompt format, the inverse scaling we identify arises in policy-relevant data and persists across model families, training stages, prompting interventions, and elicitation formats.

\begin{figure}[tb]
\centering
\includegraphics[width=\linewidth]{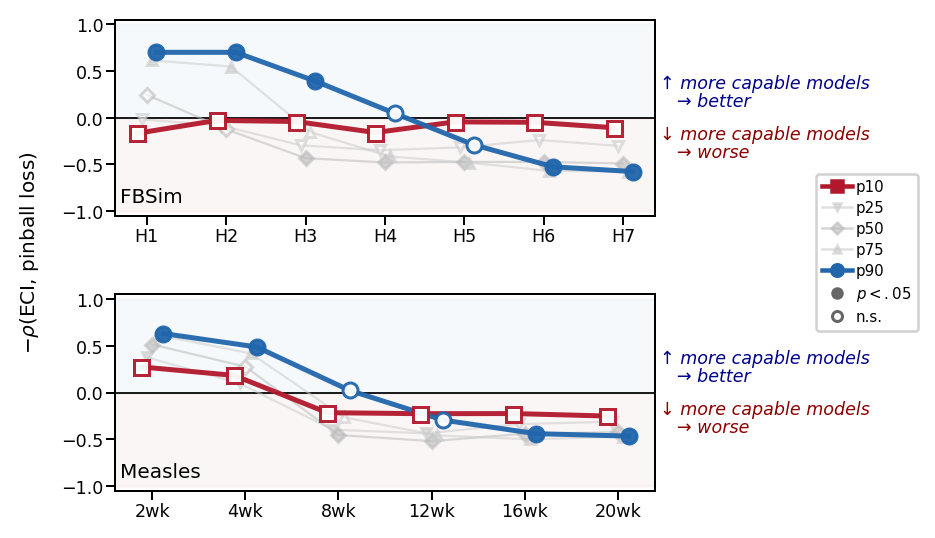}
\caption{\emph{Upper-tail predictions drive the inverse scaling, in both domains.} Per-quantile pinball-loss decomposition. \textbf{Top:} FBSim disruptable templates, $N$=28 (apples-to-apples panel; same model set as Appendix Figure~\ref{fig:horizons-hero}). \textbf{Bottom:} pre-vaccine US measles, 1928--1962, all 35~seasons pooled ($N$=28 models meeting the $\geq$80\% coverage threshold; see Appendix~\ref{sec:appendix-models}). In both domains, the p90 quantile (blue) swings from strongly positive at short horizons to strongly negative at long horizons. The p10 quantile (red) is comparatively flat. More capable models are more confident that growth will continue; the resulting upper-tail miscalibration reverses the capability--accuracy relationship under regime change.}
\label{fig:pinball}
\end{figure}

We discover this inverse scaling on FBSim, a contamination-free forecasting benchmark we release, built on procedurally generated strategy-game rollouts. Pooled across all six question templates, distributional forecast quality degrades with capability at long horizons ($\rho$=$-$0.42 at H7, $N$=28). A per-quantile decomposition reveals the source: more capable models push their upper tail upward to track growth trajectories that subsequently break, while the lower tail stays anchored (Figure~\ref{fig:pinball}).

We isolate the mechanism on a synthetic SIR epidemiological model: superlinear growth followed by regime change produces the inversion, while linear growth with the same crash structure does not. A controlled within-family Llama-3.1 $2{\times}2$ (\{70B, 405B\}$\times$\{base, instruct\}) then shows that both scale and post-training independently contribute and the two compound.

The pattern replicates on real-world time series where tail sensitivity is critical: COVID-19 incidence across 60 countries, U.S.\ housing prices during the 2003--2006 bubble, monthly CPI during 12 hyperinflationary episodes, and pre-vaccine US measles ($N$=1{,}339 state-seasons across all 35 viable seasons before vaccine licensure). Influenza, structurally similar but with insufficient overshoot, shows no inversion: the trigger is superlinear growth, not disease data per se. Per-domain Spearman $\rho$ and 95\% bootstrap CIs are reported in each section; all cross-domain CIs exclude zero. Measles is unselected on regime change and the inversion still emerges \emph{ex ante}.

Prompt interventions probe whether knowledge rescues performance. A generic uncertainty cue does not attenuate the effect on any domain; domain-specific identity has inconsistent effects: rescuing COVID-19, partially attenuating housing, SIR, and measles, and failing on hyperinflation despite models correctly identifying the crisis in 46 of 48~probes (\S\ref{sec:knowledge}).

Concerningly, this inverse scaling is \emph{invisible under binary metrics}. On FBSim, the same models that degrade under CRPS improve under Brier ($\rho$=$+$0.45, $N$=27). Derived Brier scores from the same continuous forecasts reverse the sign of the capability--performance relationship, ruling out task-format confounds. Existing LLM forecasting benchmarks \citep{karger2025forecastbench, nel2025kalshibench, halawi2024approaching} report binary or threshold metrics exclusively; on tasks with the structure we study, they would not detect the failure.

The models most likely to be deployed for tail-sensitive forecasting tasks are the ones most likely to commit prematurely to a growth trajectory, miscalibrating the upper tail when growth reverts. We recommend that probabilistic evaluations report at least one distributional scoring rule alongside threshold metrics; the within-family ablation suggests that overcommitment is unlikely to be corrected by scale alone.

\paragraph{Preliminaries.}
\label{sec:preliminaries}
We measure model capability via the Epoch Capabilities Index \citep[ECI;][]{ho2025rosetta}, an aggregate score over standard LLM benchmarks; models in our analysis span ECI $114$--$155$. Throughout, we report Spearman $\rho$ between ECI and per-model scoring-rule means, sign-adjusted so $\rho>0$ corresponds to higher-ECI models performing better (\emph{positive scaling}) and $\rho<0$ to inverse scaling \citep{mckenzie2023inverse}; for lower-is-better metrics we negate. We use three proper scoring rules: \textbf{Brier} on binary forecasts, $(p-y)^2$ with $y\in\{0,1\}$ \citep{brier1950verification}; \textbf{CRPS} on real-valued forecasts, $\text{CRPS}(F,y)=\int (F(z)-\mathbb{1}[z\geq y])^2\,dz$, equivalent to Brier integrated over all thresholds \citep{gneiting2007sharpness}; and \textbf{pinball loss}, the per-quantile component of CRPS. Distributional forecasts are elicited as five quantiles (p10, p25, p50, p75, p90) with CRPS computed from the piecewise-linear CDF. Brier scores a forecast at a single chosen threshold; CRPS sweeps Brier across the full outcome axis, so the upper-tail region of the integral picks up costs that any single salient-threshold Brier evaluation does not. For real-world and synthetic mechanism experiments (\S\S\ref{sec:mechanism}, \ref{sec:replication}, \ref{sec:measles}), reasoning-capable models are clamped to \texttt{reasoning\_effort="minimal"}; this is conservative against our hypothesis (a reasoning-on subset shows a stronger inversion, $\rho$=$-$0.56 vs $-$0.42 at H7; Appendix~\ref{sec:appendix-reasoning-mode}).

\paragraph{Identification.} ECI aggregates across capabilities that co-vary with release date, RLHF treatment, base architecture, and provider-specific alignment practices, so a cross-family $\rho$ alone cannot attribute the inverse scaling to any single axis. We address this in three ways: (a) provider fixed effects partial out provider identity (Appendix~\ref{sec:appendix-robustness}); (b) within-lineage replication isolates capability from post-training pipeline (OpenAI $N{=}10$, $p{=}.008$; Appendix~\ref{sec:appendix-robustness}); and (c) a controlled within-family $2{\times}2$ on Llama-3.1-\{70B, 405B\}$\times$\{base, instruct\} (\S\ref{sec:rlhf}) directly isolates the contributions of scale and post-training.

\section{ForecastBench-Sim}
\label{sec:civbench}

\subsection{Benchmark design}
\label{sec:benchmark-design}

ForecastBench-Sim (FBSim) is a forecasting benchmark built on rollouts from FreeCiv, an open-source turn-based empire-building game, via the CivRealm agent interface \citep{freeciv_project2026,freecivweb_project2026,qi2024civrealm}. The benchmark generates AI-vs.-AI games, freezes each at a snapshot turn, and converts the observed state into a natural-language world report (Appendix~\ref{sec:appendix-world-report}). Given this report, a model forecasts future properties of the same world at multiple horizons. The setting preserves features of real-world forecasting under uncertainty: long time horizons, partial observability, and occasional discontinuous shocks such as war, conquest, and regime change, so models can succeed by extrapolating genuine regularities but fail when overconfident that those regularities hold.

From each snapshot, FBSim generates paired binary and continuous questions. A \emph{binary} question asks whether an event or relation will hold (e.g., ``Will the Romans' treasury at turn 180 exceed their turn 60 treasury?''), scored by Brier score. The matched \emph{continuous} question asks for the future value itself, elicited as a 5-quantile distribution and scored by CRPS \citep{gneiting2007scoring}. Templates cover treasury, territory, population, city count, technology count, and overall score.

\textbf{Benchmark.} We develop FBSim in a concurrent submission \citep{anonymous2026fbsim}.

\textbf{Reproduction.} The generation pipeline, evaluation harness, and scoring code are included in supplementary material. Rollouts are procedurally generated and contamination-free relative to public training corpora.

\subsection{Results}
\label{sec:civbench-results}

FBSim shows inverse scaling on long-horizon distributional forecasts. Pooled across all six templates, the cross-model rank correlation between CRPS and ECI flips from $\rho$=$+$0.67 at H1 to $\rho$=$-$0.42 at H7 (95\% bootstrap CI [$-$0.72, $-$0.02], $N$=28; per-horizon evolution in Appendix Figure~\ref{fig:horizons-hero}, per-horizon and per-template details in Appendix~\ref{sec:appendix-disruptable}). The H7 cross-section appears in Figure~\ref{fig:hero} (top). The pooled-panel correlation we report is the unconditional, ex-ante quantity: it averages across all templates. The sign change at long horizons is what a forecaster choosing among models on the basis of headline capability scores would observe. The inversion is robust to leave-one-provider-out and one-model-per-release-lineage collapse (Appendix~\ref{sec:appendix-robustness}).

A natural question is whether the inversion is uniform across templates or localized to a structural subset. We split templates by \emph{disruptability}: a template is disruptable if decline is a modal outcome of the underlying world snapshots, computed before any per-template ECI--CRPS correlation is examined (Appendix~\ref{sec:appendix-disruptable}). Four of the six templates qualify (treasury, territory, population, city count); the two non-disruptable templates (technology count, ratcheted by construction; overall score, a bounded composite) are structurally protected from sustained decline. At H7, all four disruptable templates show negative ECI--CRPS correlations and both non-disruptable templates show positive correlations; under sign exchangeability this exact $4{-}2$ partition has probability $1/64{=}0.016$.

A per-quantile pinball decomposition on the disruptable templates (Figure~\ref{fig:pinball}) localizes the failure: the p90 quantile swings from $\rho$=$+$0.78 ($p$$<$$0.001$) at H1 to $\rho$=$-$0.57 ($p$=$0.002$) at H7, while p10 stays flat. More capable models are not worse at imagining crashes; their upper tail shifts upward to track the extrapolated trajectory while the lower tail stays anchored; when the trajectory breaks, the elevated $p_{90}$ sits far above the outcome and the pinball penalty dominates CRPS (distributions widen asymmetrically; Appendix~\ref{sec:appendix-robustness}). The pooled effect is robust to provider controls, within-lineage replication (OpenAI and Anthropic independently), template exclusion, and restricting to models with default reasoning engaged (the last strengthens H7 to $\rho$=$-$0.56; Appendices~\ref{sec:appendix-robustness} and~\ref{sec:appendix-reasoning-mode}).

\section{Isolating the mechanism}
\label{sec:mechanism}

\begin{figure}[tb]
\centering
\includegraphics[width=\linewidth]{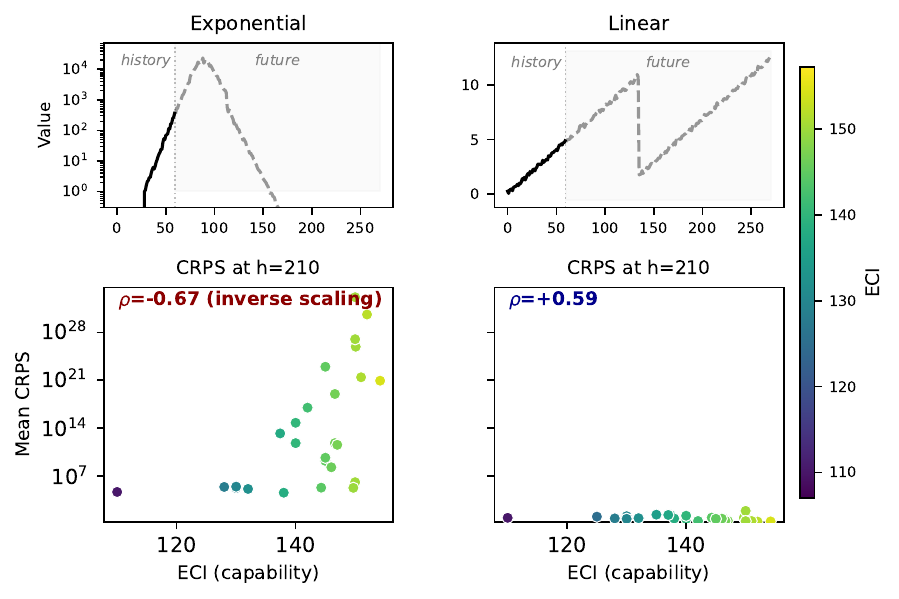}
\caption{\emph{Time series with superlinear growth and tail-risk of regime change trigger the inverse scaling.} \textbf{Top:} Ground-truth series shown to models as history (black) with continuations not shown (gray). \emph{Left}: SIR epidemic (log scale): exponential growth then intervention-driven decline. \emph{Right}: linear growth with the same downward-jump structure. \textbf{Bottom:} CRPS vs.\ ECI at $h$=210. On SIR data, more capable models produce worse distributional forecasts (inverse scaling). On linear data, the usual capability advantage holds, even when linear models continue to grow post-crash (positive scaling).}
\label{fig:mechanism}
\end{figure}


The disruptable templates share two structural properties: superlinear growth and possible regime change. FBSim cannot tell us which of these drives the inversion, and the strategy-game setting carries other potential confounds (game complexity, question format, model familiarity). To isolate the mechanism, we turn to a controlled synthetic series with the same shape.

We generate 50~series from a standard SIR (Susceptible-Infected-Recovered) epidemiological model. Each series shows exponential growth in daily new infections, followed by a public-health intervention that reduces transmission and causes cases to peak and decline (Figure ~\ref{fig:mechanism}, top). Models receive 60~steps of the rising phase as unlabeled numbers and forecast up to 210~steps ahead. CRPS shows inverse scaling at all horizons ($\rho$=$-$0.62, [$-$0.83, $-$0.26], $p$$<$0.001, $N$=27 models; Appendix~\ref{sec:appendix-lower-eci}), matching the FBSim pattern.

To confirm that superlinear growth specifically drives the inversion, we test a control with linear growth and the same downward-jump structure. The control shows positive scaling ($\rho$=$+$0.61, [$+$0.30, $+$0.82], $N$=30); the SIR and linear-control CIs are non-overlapping, so the inversion is statistically distinguishable from a generic crash effect. Crashes alone do not produce inverse scaling, even when the regime change is permanent rather than transient (Appendix~\ref{sec:appendix-regime-long}). We conclude that the effect is the combination of superlinear growth and regime change.

The mechanism is visible in the forecasts (Figure~\ref{fig:mechanism}). More capable models detect the exponential trend and extrapolate it more aggressively: the upper tail of their predictive distribution shifts upward with the growth trajectory while the lower tail stays anchored. When the intervention hits and cases decline, the elevated upper tail sits far above the post-crash outcome and errors grow superlinearly with $h$. On linear series, extrapolation errors stay bounded regardless of capability, and the usual scaling advantage holds.

\section{Isolating scale and post-training in a controlled family}
\label{sec:rlhf}

The cross-family ECI panel of \S\S\ref{sec:civbench}--\ref{sec:mechanism} aggregates over models that vary simultaneously in scale, post-training pipeline, base architecture, and training data. To isolate the contributions of scale and post-training, we run a within-family $2{\times}2$ on Llama-3.1-\{70B, 405B\}~$\times$~\{base, instruct\}, paired across 100 fresh SIR and 100 linear-control series at $h\in\{16, 90, 210\}$ with Wilcoxon signed-rank tests on CRPS ratios. The base/instruct contrast bundles RLHF with broader post-training. Methodology details (rationale, FP8 precision matching, elicitation, per-horizon tables) are in Appendix~\ref{sec:appendix-rlhf}.

At $h{=}210$ on the crash regime, scale and post-training each significantly amplify CRPS within their respective marginal, and they compound (interaction $p_W{<}.0001$, Table~\ref{tab:rlhf-2x2}). The locus of post-training damage propagates with scale: at 70B it is a tail effect that spares the typical crash series, and at 405B it dominates the central tendency. Concretely, the fraction of crash series receiving $\geq$10$\times$ inflation grows from 41\% at 70B to 63\% at 405B (medians and per-horizon breakdown in Table~\ref{tab:rlhf-2x2}). The linear no-crash control is null at both scales with zero series in the upper tail (oppositely-signed interaction $p_W{=}.029$).

\section{Replication on real-world data}
\label{sec:replication}


Does this inverse scaling pattern replicate on real-world data with the same structure? We test on three domains: COVID-19 daily incidence (60~countries, 60-day history; Our World in Data), monthly S\&P/Case-Shiller home prices for 19~U.S.\ metros (60-month history through December~2005), and monthly CPI from 12~hyperinflationary episodes across 10~countries (with CRPS normalized by series scale). All three are selected on regime change; the unselected measles cohort is in \S\ref{sec:measles}. With raw numbers and no domain labels, the effect replicates: COVID-19 ($\rho$=$-$0.54 [$-$0.78, $-$0.19], H1--H7 = 30$k$ days ahead, $N$=30), housing at 36~months ($\rho$=$-$0.67 [$-$0.82, $-$0.40], $N$=30), and hyperinflation at $h\in\{12,24,36,48\}$~months ($\rho$=$-$0.59 [$-$0.82, $-$0.25], $N$=30). A minimal cue (``the current trend may or may not continue'') fails to attenuate on COVID-19 ($\rho$=$-$0.60) or hyperinflation ($\rho$=$-$0.50), and partially attenuates on housing ($\rho$=$-$0.26, n.s.).

\section{Testing on a natural distribution: Measles 1928--1962}
\label{sec:measles}


We test the inversion on the full pre-vaccine era of US measles using weekly state-level case counts from Project Tycho \citep{vanpanhuis2018tycho}: 35~seasons spanning 1928--1962. We exclude 1963--1965 because the Edmonston-B measles vaccine was licensed in March~1963; later seasons reflect partial vaccine deployment rather than natural epidemic dynamics. For each state-season we apply quality filters (Appendix~\ref{sec:appendix-measles}): $\geq$~30 weeks of source data, peak case count $\geq$~50, and $\geq$~30 weeks of constructed history+future. This yields $N$=1{,}339 series across 56~states. Each series presents 12~weeks of history from a late-summer trough through the early ascending phase, followed by 20+ weeks of future through the epidemic peak and decline.

Pooled across all 35~seasons (\emph{ex-ante}), the bottom panel of Appendix Figure~\ref{fig:horizons-hero} shows the same positive-to-inverse-scaling crossover observed on FBSim and the SIR synthetic data; the long-horizon ($h{=}20$) cross-section is the bottom panel of Figure~\ref{fig:hero}. At short horizons, more capable models produce better distributional forecasts (2-week: $\rho$=$+$0.64, [$+$0.33, $+$0.85], $N$=29 models scored on 1{,}339 state-seasons). By 8~weeks the correlation crosses zero, and at 12--20~weeks CRPS shows inverse scaling ($\rho$=$-$0.42, [$-$0.70, $-$0.03] at 16~weeks; $\rho$=$-$0.41, [$-$0.69, $-$0.03] at 20~weeks). CRPS inverts while Brier stays flat. This pooled effect emerges without any pre-selection by severity or other season properties: we test every viable state-season in the pre-vaccine window.

The mechanism is identical to FBSim: the per-quantile pinball decomposition (Figure~\ref{fig:pinball}, bottom panel) shows the p90 estimate swinging from strongly positive at short horizons ($\rho$=$+$0.63 at 2~weeks) to strongly negative at long horizons ($\rho$=$-$0.44 at 16~weeks), a magnitude comparable to FBSim's p90 swing. The p10 quantile is comparatively stable in both domains. In both domains, more capable models commit higher upper-tail estimates: their elevated $p_{90}$ better tracks actual seasonal peaks, but on the routine state-seasons that fill out the cohort the gap between the elevated $p_{90}$ and the low outcome dominates the cross-season pinball loss.

\paragraph{Informative negative: influenza.} We pre-specified the trigger as superlinear growth with tail risk of regime change before testing either disease; if the effect were instead an artifact of disease data in general, it should appear on influenza too. We test this on two flu datasets: modern CDC ILINet surveillance (50~series, 10~HHS regions $\times$ 5~seasons, $N$=$27$) and historical influenza from Project Tycho (1919--1951, explosive epidemic years only, $N$=$7$). Both show positive scaling or null effects: ILINet pooled $\rho$=$+$0.14 ($p$=$0.50$), historical flu $\rho$=$+$0.22 ($p$=$0.64$). Even the most explosive historical flu epidemics produce a median overshoot of only $3\times$; well below that of the most severe measles outbreaks in this corpus.

\paragraph{Domain class disclosure attenuates the inversion.}
A natural objection to the preceding result is that models did not know what they were forecasting. We test this on the same cohort with a single sentence of minimum-viable domain identification prepended to each prompt: \emph{``This time series represents the trajectory of a communicable disease in a population over time.''} The cue names the data type but does not name the disease, state, or year, and does not leak dynamic information about regime change or interventions. Restricting to the $N{=}23$ models with $\geq$95\% coverage on the cued stratum, the cue attenuates the inversion at long horizons: $\rho$=$-$0.49$\to$$-$0.39 ($p$=$.07$) at $h{=}12$wk, $-$0.45$\to$$-$0.27 ($p$=$.21$) at $h{=}16$wk, $-$0.50$\to$$-$0.08 ($p$=$.73$) at $h{=}20$wk. Domain knowledge partially rescues calibration on this naturalistic cohort, but the rescue is incomplete: capable models that recognize the data type continue to over-commit on the most severe state-seasons even as their average performance improves on the calm majority. The pattern parallels the inconsistent named-context findings of \S\ref{sec:knowledge}: more information about the domain helps where the prior fits and fails where the upper tail materializes.

\begin{figure}[tb]
\centering
\includegraphics[width=\linewidth]{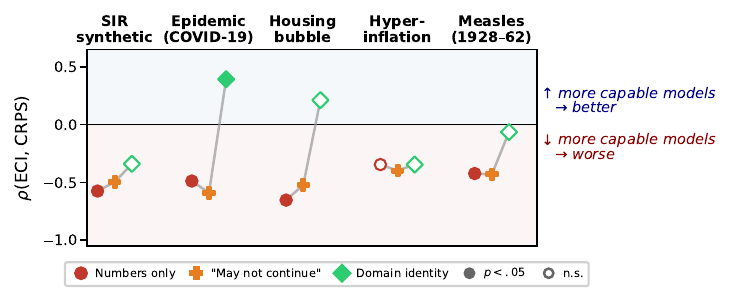}
\caption{\emph{Domain knowledge has inconsistent effects across domains.} Naming the domain rescues positive scaling on COVID-19 ($\rho$: $-$0.49~$\to$~$+$0.39), substantially attenuates it on housing ($\Delta\rho$=$+$0.86), measles ($\Delta\rho$=$+$0.36), and SIR ($\Delta\rho$=$+$0.24), but has essentially no effect on hyperinflation ($\Delta\rho$=$+$0.00). Red dots: unlabeled numbers (inverse-scaling baseline). Orange crosses: ``the current trend may or may not continue.'' Green diamonds: domain-specific identity.}
\label{fig:context}
\end{figure}

\section{Domain knowledge has inconsistent effects}
\label{sec:knowledge}


Our real-world replication (\S\ref{sec:replication}) tested raw numeric series. Presumably, identifying the domain (e.g., ``these are S\&P/Case-Shiller HPI values for [city]'') would eliminate the inverse-scaling trend: more capable models should be better able to recall and apply domain knowledge. Surprisingly, domain identity does not produce a uniform effect (Figure~\ref{fig:context}). Naming the COVID-19 country and start date crosses to significant positive scaling; on housing the correlation crosses zero; SIR and measles attenuate substantially, with measles attenuating even under minimum-viable disclosure (\S\ref{sec:measles}); on hyperinflation no attenuation occurs.

This is not a knowledge gap. Asked directly about the hyperinflation episodes, models correctly identify the inflationary crisis in 46 of 48~probes (Appendix~\ref{sec:appendix-knowledge-probe}). The relevant priors are recoverable from the models' own representations, but those priors do not reliably translate into calibrated tails.

\paragraph{During-task articulation.} Beyond the post-hoc probe, the recoverable prior surfaces during forecasting itself. Opus-4-6, forecasting the 1985--1989 Brazilian hyperinflation series under minimum-viable domain disclosure (\S\ref{sec:measles}), wrote: ``\emph{hyperinflation could also stabilize (currency reform), adding downside uncertainty. But following the trend\ldots}'' before producing $p_{50}$ percentiles roughly seven million times above the eventual outcome. The regime-change alternative is recalled, weighed, and discarded in the same response.

Reconciling the contrast across domains (naming rescues COVID-19 but leaves hyperinflation untouched) requires probing internal representations, which we treat in a separate paper.

\section{Single-threshold scoring misses the upper-tail failure}
\label{sec:threshold}

Scored using CRPS, we find a strong, persistent inverse-scaling trend across FBSim, the synthetic SIR mechanism, and real-world data. FBSim includes paired binary and continuous questions on the same worlds: on the natively binary questions, more capable models forecast better ($\rho$=$+$0.45, $N$=27), consistent with prior work on LLM forecasting \citep{karger2025forecastbench}; on the matched continuous questions scored by CRPS, the relationship reverses at long horizons ($\rho$=$-$0.42 at H7). Same models, same worlds, opposite verdicts. The result is not a question-format artifact: deriving Brier from the same five-quantile continuous forecasts reproduces the positive-scaling sign on the disruptable panel ($\rho$=$+$0.50, $p$=$0.007$, $N$=28; threshold construction in Appendix~\ref{sec:appendix-robustness}). Identical outputs, opposite scaling sign under different scoring rules; the cost lies in the upper-tail integral, which CRPS captures and a single-threshold Brier does not.

\citet{schaeffer2023mirage} showed that metric choice can manufacture apparent emergence from smooth capability trends. Our finding is a different and arguably more concerning version of the same problem: metric choice does not merely inflate or deflate the trend, it \emph{reverses the sign}. A benchmark that reports only Brier or accuracy would certify that more capable models are better forecasters on our data. Existing LLM forecasting evaluations (ForecastBench \citep{karger2025forecastbench}, KalshiBench \citep{nel2025kalshibench}, and others \citep{halawi2024approaching, schoenegger2024wisdom}) report binary or threshold metrics exclusively. On tasks with the structure we study, these evaluations would not detect the distributional failure.

\section{Discussion}
\label{sec:discussion}

Our central finding is methodological as much as empirical: single-threshold scoring at the conventional cutoff reverses the sign of capability scaling relative to tail-inclusive scoring on the same outputs. Brier at the salient threshold is mathematically a component of CRPS; what conventional evaluations omit is the upper-tail region where elevated $p_{90}$ accumulates cost. We demonstrate this reversal across our procedurally generated benchmark (FBSim), a synthetic SIR mechanism, three real-world domains (COVID-19, housing, hyperinflation), and the natural distribution of 35 pre-vaccine US measles seasons. The FBSim and the measles findings are both \emph{ex ante}, pooled across all templates and all 35 pre-vaccine seasons before any structural decomposition.

The inverse-scaling pattern we document is distinct from those documented in existing literature. It is not distraction by misleading surface features \citep{mckenzie2023inverse}, and it is not a U-shaped pattern that resolves at frontier scale \citep{wei2023ushaped}: the most capable models sit at the worst end. Instead, more capable models detect growth signals more precisely and commit to them more strongly by shifting the upper tail upward, improving $p_{90}$ calibration on cells where outcomes reach the upper tail and paying the cost on the more common cells where they stay near the median: competence-driven overcommitment rather than error. The same mechanism explains why prompt interventions and domain naming fail to rescue calibration on most domains (\S\ref{sec:knowledge}): the relevant priors are recoverable from the models' own representations (models correctly identify the hyperinflation crises in 46 of 48 probes, and sometimes articulate the regime-change alternative mid-forecast before discarding it) but those priors do not reliably translate into calibrated tails.

The domains where distributional accuracy matters most, epidemics (when to intervene), monetary policy (tail inflation risk), and financial risk management (value-at-risk, expected shortfall), are exactly the domains with the superlinear-growth-plus-regime-change structure that triggers this failure. In these settings, the modal forecast is the wrong thing to optimize: the tails determine whether a hospital system is overwhelmed, a currency collapses, or a portfolio suffers a margin call. Disease surveillance is the most immediate case we document: public-health agencies forecast case trajectories every season to time interventions and allocate resources, active research now applies LLMs to this task \citep{du2025pandemicllm} and, at the time of writing, measles is a leading global health concern amid resurgent outbreaks. Our results suggest that LLM-driven epidemic forecasts will systematically miscalibrate the upper tail under regime change --- overshooting on routine seasons and missing the timing of actual peaks --- in domains where a mis-timed forecast becomes preventable hospitalizations and deaths.

Current evaluation and training infrastructure both may contribute to failure. In evaluation, single-threshold scoring at conventional salient cutoffs (common in LLM forecasting benchmarks \citep{karger2025forecastbench, nel2025kalshibench, halawi2024approaching, schoenegger2024wisdom}) misses the upper-tail cost entirely, certifying capability where tail-inclusive scoring on the same outputs reveals degradation. In training, post-training may compound the issue: the within-family $2{\times}2$ in \S\ref{sec:rlhf} shows that the base$\to$instruct shift significantly amplifies the long-horizon crash-regime failure within Llama-3.1, with pure scale aggravating it independently.

We recommend that LLM forecasting evaluations probe the tails: either via a tail-integrating proper rule (CRPS, log score) or via Brier swept across upper-tail thresholds. A single-threshold Brier at the conventional salient cutoff is not enough; on tasks with the structure we study, it certifies capability where tail-inclusive scoring on the same outputs reveals failure. If training objectives continue to reward threshold accuracy, the overcommitment we document is unlikely to be corrected at scale and may compound.

\subsection{Limitations}
\label{sec:limitations}
The paper's claim is bounded by design and by available data.

\textbf{Scope.} We characterize a structurally identifiable class of inverse scaling (time series with superlinear growth and tail risk of regime change) and do not claim this is the only kind of inverse scaling, nor that it appears uniformly across forecasting tasks.
\textbf{Domain selection.} Three of our four real-world replication domains (housing, hyperinflation, COVID-19) were chosen because the regime change the mechanism predicts had already occurred: a precondition for testing whether the inversion appears, but one that prevents us from claiming how often models fail this way in deployment. The 35-season pre-vaccine measles cohort, which we do not pre-select on severity, anchors the ex-ante view (\S\ref{sec:measles}).
\textbf{Capability is mostly observed, not manipulated.} On the cross-family panel we measure capability with an external index (ECI) rather than varying it directly, because not every capability axis can be controlled across model families. We address this with within-lineage replication and provider fixed effects (Appendix~\ref{sec:appendix-robustness}), and a partition by reasoning configuration at evaluation time (Appendix~\ref{sec:appendix-reasoning-mode}). The within-family Llama-3.1 $2{\times}2$ in \S\ref{sec:rlhf} additionally manipulates two capability axes directly (scale: 70B vs 405B; and alignment-stage training: base vs instruct), and the inversion follows both, so the cross-family ECI gradient is not the only manipulation backing the result. Alternative composite capability orderings remain future work.
\textbf{Sample size.} Our hyperinflation ($N$=12 episodes) and housing ($N$=19 metros) samples are bounded by the historical record, which makes the cross-model Spearman~$\rho$ noisier on those domains; we report them as directional findings and rely on the synthetic SIR (\S\ref{sec:mechanism}) and 35-season measles (\S\ref{sec:measles}) cohorts for precise effect-size estimates. We address generic-domain prompts on the measles cohort: minimum-viable domain disclosure attenuates the inversion (\S\ref{sec:measles}), partially extending the inconsistent named-context findings of \S\ref{sec:knowledge}. Generic-domain prompts on housing, hyperinflation, and COVID-19 against their true unconditional distributions are left as future work.
\textbf{Mechanistic interpretability.} The most surprising result in this paper is the knowledge--calibration gap in \S\ref{sec:knowledge}: naming the domain attenuates but does not consistently rescue inverse scaling, even though models can recall the relevant priors when asked directly (46 of 48 hyperinflation probes correctly identify the crisis). Prompt-level interventions cannot tell us why a recoverable prior fails to translate into calibrated tails. We treat the mechanistic question (probing internal representations of regime, trend, and the elicited quantile) in a separate paper.

\section{Related Work}
\label{sec:related-work}

\textbf{Inverse scaling.} \citet{mckenzie2023inverse} document inverse scaling on tasks where larger models are more susceptible to distraction or sycophancy; \citet{wei2023ushaped} show many such cases U-shape with further scaling. Our pattern is distinct: it has not resolved at frontier scale, and the mechanism is overcommitment, not distraction. \citet{schaeffer2023mirage} show that nonlinear metrics can manufacture apparent emergence from smooth trends; we show that proper scoring-rule choice can reverse the \emph{direction} of scaling on identical outputs.

\textbf{LLM forecasting and post-training.} Prior work establishes capability-to-binary-accuracy on forecasting \citep{halawi2024approaching, karger2025forecastbench, nel2025kalshibench, ye2024miraievaluatingllmagents, wildman2025pastcasting}; we replicate under Brier and show reversal under CRPS on identical outputs. Closest on post-training, \citet{gruver2023llmtime} report alignment-tuned LLaMA-2 chat under-performing base under CRPS on Darts; our $2{\times}2$ (\S\ref{sec:rlhf}) extends this to scale$\times$post-training.

\bibliographystyle{plainnat}
\bibliography{references}

\newpage
\appendix

\section{Models}
\label{sec:appendix-models}

We evaluate 29 models spanning ECI 114--155 from seven providers. Table~\ref{tab:models} lists all models with their capability scores. Three models in the Epoch Capabilities Index were excluded from all analyses: \textbf{Mixtral-8x7B} was screened out before data collection (0\% parse success on a pilot batch of quantile-elicitation prompts; no raw FBSim data collected); \textbf{claude-3-haiku} was attempted at full collection but produced 0\% parseable forecasts; \textbf{gemini-2.0-flash-lite} was attempted at full collection but exceeded a 30\% parse-failure threshold.

\begin{table}[h]
\centering
\small
\caption{Models evaluated, ordered by ECI score (Epoch AI, accessed 2026-04-30).}
\label{tab:models}
\begin{tabular}{llr}
\toprule
Provider & Model & ECI \\
\midrule
OpenAI & gpt-3.5-turbo-0125 & 114 \\
Meta/Together & Llama-3.3-70B-Instruct-Turbo & 128 \\
Mistral & mistral-large-2407 & 128 \\
OpenAI & gpt-4o & 129 \\
Mistral & mistral-large-2411 & 129 \\
Mistral & mistral-large-latest & 129 \\
DeepSeek/Together & DeepSeek-V3 & 137 \\
OpenAI & gpt-4.1-2025-04-14 & 138 \\
DeepSeek/Together & DeepSeek-V3.1 & 139 \\
OpenAI & gpt-5-nano-2025-08-07 & 141 \\
OpenAI & o3-mini-2025-01-31 & 142 \\
Anthropic & claude-haiku-4-5-20251001 & 143 \\
Anthropic & claude-sonnet-4-20250514 & 143 \\
Google & gemini-2.5-flash & 143 \\
xAI & grok-4-fast-non-reasoning & 145 \\
xAI & grok-4-fast-reasoning & 145 \\
OpenAI & gpt-5-mini-2025-08-07 & 146 \\
xAI & grok-4-1-fast-non-reasoning & 146 \\
Anthropic & claude-sonnet-4-5-20250929 & 147 \\
OpenAI & o3-2025-04-16 & 147 \\
OpenAI & o4-mini-2025-04-16 & 147 \\
Google & gemini-2.5-pro & 147 \\
xAI & grok-4-0709 & 147 \\
Anthropic & claude-opus-4-5-20251101 & 150 \\
OpenAI & gpt-5-2025-08-07 & 150 \\
OpenAI & gpt-5.1-2025-11-13 & 150 \\
xAI & grok-4-1-fast-reasoning & 151 \\
Google & gemini-3-pro-preview & 153 \\
Anthropic & claude-opus-4-6 & 155 \\
\bottomrule
\end{tabular}
\end{table}

FBSim raw data was collected for 31 models (the 29 in Table~\ref{tab:models} plus claude-3-haiku and gemini-2.0-flash-lite; Mixtral-8x7B was screened out before data collection and produced no raw forecasts). The two collected-then-excluded models are filtered out of every continuous-CRPS analysis by the parse-failure exclusions described above and do not enter any reported $\rho$. Time-series evaluations draw from the 29 models listed above.

\begin{figure}[!h]
\centering
\includegraphics[width=\linewidth]{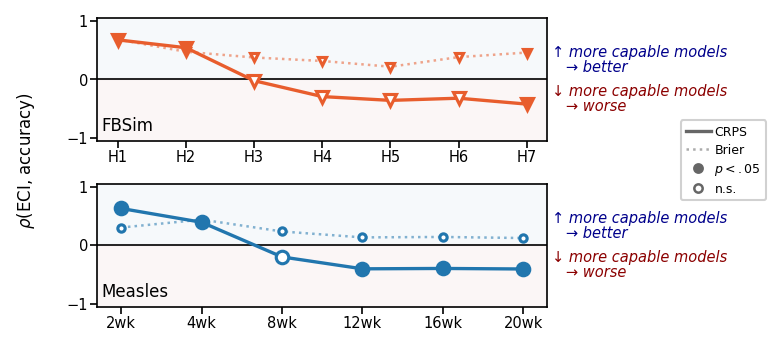}
\caption{\emph{Across-horizon evolution of the capability--accuracy relationship.} Spearman $\rho$ between model capability (Epoch Capabilities Index) and forecast accuracy vs.\ horizon, sign-flipped so positive = positive scaling, negative = inverse scaling. \textbf{Top:} FBSim, pooled across all six question templates (H1--H7 = game turns). \textbf{Bottom:} pre-vaccine US measles case counts, 1928--1962, pooled across all 35 seasons. In both domains, tail-inclusive scoring (solid: CRPS) shows the positive-to-inverse-scaling crossover; single-threshold Brier at the salient cutoff (dotted) does not. Both panels share the $N{=}28$ model panel described above; 95\% bootstrap CIs are reported per-horizon in the body and \S\ref{sec:appendix-robustness}.}
\label{fig:horizons-hero}
\end{figure}

\paragraph{Cross-domain figures.} The headline results in \S\ref{sec:civbench-results} and Figures~\ref{fig:horizons-hero} and~\ref{fig:pinball} use the panel of $N$=28 models with both continuous FBSim coverage and Rule-A coverage of pre-vaccine measles, so the FBSim and measles correlations are computed on the same model set. The single model excluded from the panel by Rule A is grok-4-0709 (19\% measles coverage; see Table~\ref{tab:availability}). The cross-section in Figure~\ref{fig:hero} drops a further five measles models whose mean CRPS at $h{=}20$ exceeded 100K cases (cohort match across panels); Figure~\ref{fig:horizons-hero} shows the corresponding across-horizon evolution on the full $N$=28 cohort.

\paragraph{Per-stratum inclusion criterion.} For each time-series stratum, a model is included in the cross-model correlation only if it has successfully scored at least 80\% of that stratum's series. Models below this coverage threshold (typically due to systematic parse failures on a stratum's prompt format, or to incomplete runs) are excluded from that stratum's analysis but retained for any other stratum where they meet the threshold. This rule applies uniformly across all 19 time-series strata and replaces ad-hoc per-stratum exclusion notes. Per-stratum $N$ and individual exclusions appear in Table~\ref{tab:availability}.

\begin{sidewaystable}[p]
\centering\tiny
\setlength{\tabcolsep}{2.8pt}
\caption{\emph{Model $\times$ stratum availability under Rule A.} Cell entries: $\checkmark$ = $\geq$80\% coverage (included in stratum's analysis); two-digit number = coverage \% below threshold (excluded); $-$ = no run. Bottom row gives per-stratum $N$.}
\label{tab:availability}
\begin{tabular}{lr|ccccccccccccccccccc}
\toprule
 & & \multicolumn{3}{c}{\textit{Discovery}} & \multicolumn{3}{c}{\textit{Replication}} & \multicolumn{3}{c}{\textit{Natural}} & \multicolumn{10}{c}{\textit{Knowledge interventions}} \\
Model & ECI & \rotatebox{60}{SIR} & \rotatebox{60}{SIRl} & \rotatebox{60}{SIRl*} & \rotatebox{60}{EPI} & \rotatebox{60}{HOUS} & \rotatebox{60}{HYP} & \rotatebox{60}{MEAS} & \rotatebox{60}{FLU} & \rotatebox{60}{FLUh} & \rotatebox{60}{SIRn} & \rotatebox{60}{SIRp} & \rotatebox{60}{EPIc} & \rotatebox{60}{EPIn} & \rotatebox{60}{HOUSc} & \rotatebox{60}{HOUSn} & \rotatebox{60}{HYPc} & \rotatebox{60}{HYPn} & \rotatebox{60}{MEASc} & \rotatebox{60}{MEASn} \\
\midrule
gpt-3.5-turbo-0125 & 114.0 & $\checkmark$ & $\checkmark$ & $\checkmark$ & $\checkmark$ & $\checkmark$ & $\checkmark$ & $\checkmark$ & $\checkmark$ & $\checkmark$ & $\checkmark$ & $\checkmark$ & $\checkmark$ & $\checkmark$ & $\checkmark$ & $\checkmark$ & $\checkmark$ & $\checkmark$ & $\checkmark$ & $\checkmark$ \\
Llama-3.3-70B & 128.0 & $-$ & $\checkmark$ & $\checkmark$ & $\checkmark$ & $\checkmark$ & \scriptsize{75} & $\checkmark$ & $\checkmark$ & $-$ & $\checkmark$ & $-$ & $\checkmark$ & $\checkmark$ & $\checkmark$ & \scriptsize{53} & $\checkmark$ & \scriptsize{67} & $\checkmark$ & $\checkmark$ \\
mistral-large-2407 & 128.0 & $\checkmark$ & $\checkmark$ & $\checkmark$ & $\checkmark$ & $\checkmark$ & $\checkmark$ & $\checkmark$ & $\checkmark$ & $-$ & $\checkmark$ & $\checkmark$ & \scriptsize{72} & $\checkmark$ & $\checkmark$ & $\checkmark$ & $\checkmark$ & $\checkmark$ & $\checkmark$ & $\checkmark$ \\
gpt-4o & 129.0 & $\checkmark$ & $\checkmark$ & $\checkmark$ & $\checkmark$ & $\checkmark$ & $\checkmark$ & $\checkmark$ & $\checkmark$ & $\checkmark$ & $\checkmark$ & $\checkmark$ & $\checkmark$ & $\checkmark$ & $\checkmark$ & \scriptsize{5} & $\checkmark$ & $\checkmark$ & $\checkmark$ & $\checkmark$ \\
mistral-large-2411 & 129.0 & $\checkmark$ & $\checkmark$ & $\checkmark$ & $\checkmark$ & $\checkmark$ & \scriptsize{58} & $\checkmark$ & $\checkmark$ & $-$ & $\checkmark$ & $\checkmark$ & $\checkmark$ & $\checkmark$ & $\checkmark$ & $\checkmark$ & $\checkmark$ & $\checkmark$ & $\checkmark$ & \scriptsize{36} \\
mistral-large-latest & 129.0 & $\checkmark$ & $\checkmark$ & $\checkmark$ & $\checkmark$ & $\checkmark$ & $\checkmark$ & $\checkmark$ & $\checkmark$ & $-$ & \scriptsize{78} & $\checkmark$ & \scriptsize{78} & $\checkmark$ & $\checkmark$ & $\checkmark$ & $\checkmark$ & $\checkmark$ & $\checkmark$ & $\checkmark$ \\
DeepSeek-V3 & 137.0 & $-$ & $\checkmark$ & $\checkmark$ & $\checkmark$ & $\checkmark$ & \scriptsize{75} & $\checkmark$ & $\checkmark$ & $-$ & $\checkmark$ & $-$ & $\checkmark$ & $\checkmark$ & $\checkmark$ & \scriptsize{63} & $\checkmark$ & $\checkmark$ & $\checkmark$ & $\checkmark$ \\
gpt-4.1-25-04-14 & 138.0 & $\checkmark$ & $\checkmark$ & $\checkmark$ & $\checkmark$ & \scriptsize{79} & $\checkmark$ & $\checkmark$ & $\checkmark$ & $-$ & $\checkmark$ & $\checkmark$ & $\checkmark$ & $\checkmark$ & $\checkmark$ & $\checkmark$ & $\checkmark$ & \scriptsize{8} & $\checkmark$ & $\checkmark$ \\
DeepSeek-V3.1 & 139.0 & $-$ & $\checkmark$ & $\checkmark$ & $\checkmark$ & $\checkmark$ & $\checkmark$ & $\checkmark$ & $\checkmark$ & $-$ & $\checkmark$ & $-$ & $\checkmark$ & $\checkmark$ & $\checkmark$ & $\checkmark$ & $\checkmark$ & \scriptsize{75} & $\checkmark$ & $\checkmark$ \\
gpt-5-nano-25-08-07 & 141.0 & $\checkmark$ & $\checkmark$ & $\checkmark$ & $\checkmark$ & $\checkmark$ & $\checkmark$ & $\checkmark$ & $\checkmark$ & $-$ & $\checkmark$ & $\checkmark$ & $\checkmark$ & $\checkmark$ & $\checkmark$ & \scriptsize{74} & $\checkmark$ & $\checkmark$ & $\checkmark$ & $\checkmark$ \\
o3-mini-25-01-31 & 142.0 & $\checkmark$ & $\checkmark$ & $\checkmark$ & $\checkmark$ & $\checkmark$ & $\checkmark$ & $\checkmark$ & $\checkmark$ & $-$ & $\checkmark$ & $\checkmark$ & $\checkmark$ & $\checkmark$ & $\checkmark$ & $\checkmark$ & $\checkmark$ & $\checkmark$ & $\checkmark$ & $\checkmark$ \\
claude-haiku-4-5 & 143.0 & $\checkmark$ & $\checkmark$ & $\checkmark$ & $\checkmark$ & $\checkmark$ & $\checkmark$ & $\checkmark$ & $\checkmark$ & $\checkmark$ & $\checkmark$ & $\checkmark$ & $\checkmark$ & $\checkmark$ & $\checkmark$ & $\checkmark$ & $\checkmark$ & $\checkmark$ & $\checkmark$ & $\checkmark$ \\
claude-sonnet-4 & 143.0 & $\checkmark$ & $\checkmark$ & $\checkmark$ & $\checkmark$ & $\checkmark$ & $\checkmark$ & $\checkmark$ & $\checkmark$ & $-$ & $\checkmark$ & $\checkmark$ & $\checkmark$ & $\checkmark$ & $\checkmark$ & $\checkmark$ & $\checkmark$ & $\checkmark$ & $\checkmark$ & $\checkmark$ \\
gemini-2.5-flash & 143.0 & $\checkmark$ & $\checkmark$ & $\checkmark$ & $\checkmark$ & $\checkmark$ & $\checkmark$ & $\checkmark$ & $\checkmark$ & $\checkmark$ & $\checkmark$ & $\checkmark$ & $\checkmark$ & $\checkmark$ & $\checkmark$ & $\checkmark$ & $\checkmark$ & $\checkmark$ & $\checkmark$ & $\checkmark$ \\
grok-4-fast-non-reasoning & 145.0 & $\checkmark$ & $\checkmark$ & $\checkmark$ & $\checkmark$ & $\checkmark$ & $\checkmark$ & $\checkmark$ & $\checkmark$ & $-$ & $\checkmark$ & $\checkmark$ & $\checkmark$ & $\checkmark$ & $\checkmark$ & $\checkmark$ & $\checkmark$ & $\checkmark$ & $\checkmark$ & $\checkmark$ \\
grok-4-fast-reasoning & 145.0 & $\checkmark$ & $\checkmark$ & $\checkmark$ & $\checkmark$ & $\checkmark$ & $\checkmark$ & $\checkmark$ & $\checkmark$ & $-$ & $\checkmark$ & $\checkmark$ & $\checkmark$ & $\checkmark$ & $\checkmark$ & $\checkmark$ & $\checkmark$ & $\checkmark$ & $\checkmark$ & $\checkmark$ \\
gpt-5-mini-25-08-07 & 146.0 & $\checkmark$ & $\checkmark$ & $\checkmark$ & $\checkmark$ & $\checkmark$ & $\checkmark$ & $\checkmark$ & $\checkmark$ & $-$ & $\checkmark$ & $\checkmark$ & $\checkmark$ & $\checkmark$ & $\checkmark$ & $\checkmark$ & $\checkmark$ & $\checkmark$ & $\checkmark$ & $\checkmark$ \\
grok-4-1-fast-non-reasoning & 146.0 & $\checkmark$ & $\checkmark$ & $\checkmark$ & $\checkmark$ & $\checkmark$ & $\checkmark$ & $\checkmark$ & $\checkmark$ & $-$ & $\checkmark$ & $\checkmark$ & $\checkmark$ & $\checkmark$ & $\checkmark$ & $\checkmark$ & $\checkmark$ & $\checkmark$ & $\checkmark$ & $\checkmark$ \\
claude-sonnet-4-5 & 147.0 & $\checkmark$ & $\checkmark$ & $\checkmark$ & $\checkmark$ & $\checkmark$ & $\checkmark$ & $\checkmark$ & $\checkmark$ & $\checkmark$ & $\checkmark$ & $\checkmark$ & $\checkmark$ & $\checkmark$ & $\checkmark$ & $\checkmark$ & $\checkmark$ & $\checkmark$ & $\checkmark$ & $\checkmark$ \\
gemini-2.5-pro & 147.0 & $\checkmark$ & $\checkmark$ & $\checkmark$ & $\checkmark$ & $\checkmark$ & $\checkmark$ & $\checkmark$ & $\checkmark$ & $-$ & $\checkmark$ & $\checkmark$ & $\checkmark$ & $\checkmark$ & $\checkmark$ & $\checkmark$ & $\checkmark$ & $\checkmark$ & $\checkmark$ & $\checkmark$ \\
grok-4-0709 & 147.0 & $\checkmark$ & $\checkmark$ & $\checkmark$ & $\checkmark$ & $\checkmark$ & $\checkmark$ & \scriptsize{19} & $\checkmark$ & $-$ & $\checkmark$ & $\checkmark$ & $\checkmark$ & $\checkmark$ & $\checkmark$ & $\checkmark$ & $\checkmark$ & $\checkmark$ & \scriptsize{16} & $-$ \\
o3-25-04-16 & 147.0 & $\checkmark$ & $\checkmark$ & $\checkmark$ & $\checkmark$ & $\checkmark$ & $\checkmark$ & $\checkmark$ & $\checkmark$ & $-$ & $\checkmark$ & $\checkmark$ & $\checkmark$ & $\checkmark$ & $\checkmark$ & $\checkmark$ & $\checkmark$ & $\checkmark$ & $\checkmark$ & $\checkmark$ \\
o4-mini-25-04-16 & 147.0 & $\checkmark$ & $\checkmark$ & $\checkmark$ & $\checkmark$ & $\checkmark$ & $\checkmark$ & $\checkmark$ & $\checkmark$ & $-$ & $\checkmark$ & $\checkmark$ & $\checkmark$ & $\checkmark$ & $\checkmark$ & $\checkmark$ & $\checkmark$ & $\checkmark$ & $\checkmark$ & $\checkmark$ \\
claude-opus-4-5 & 150.0 & $\checkmark$ & $\checkmark$ & $-$ & $\checkmark$ & $\checkmark$ & $\checkmark$ & $\checkmark$ & $-$ & $-$ & $\checkmark$ & $\checkmark$ & $\checkmark$ & $\checkmark$ & $\checkmark$ & $\checkmark$ & $\checkmark$ & $\checkmark$ & $\checkmark$ & $\checkmark$ \\
gpt-5-25-08-07 & 150.0 & $\checkmark$ & $\checkmark$ & $\checkmark$ & $\checkmark$ & $\checkmark$ & $\checkmark$ & $\checkmark$ & $\checkmark$ & $\checkmark$ & $\checkmark$ & $\checkmark$ & $\checkmark$ & $\checkmark$ & $\checkmark$ & $\checkmark$ & $\checkmark$ & $\checkmark$ & $\checkmark$ & $\checkmark$ \\
gpt-5.1-25-11-13 & 150.0 & $\checkmark$ & $\checkmark$ & $\checkmark$ & $\checkmark$ & $\checkmark$ & $\checkmark$ & $\checkmark$ & $\checkmark$ & $\checkmark$ & $\checkmark$ & $\checkmark$ & $\checkmark$ & $\checkmark$ & $\checkmark$ & $\checkmark$ & $\checkmark$ & $\checkmark$ & $\checkmark$ & $\checkmark$ \\
grok-4-1-fast-reasoning & 151.0 & $\checkmark$ & $\checkmark$ & $\checkmark$ & \scriptsize{77} & $\checkmark$ & $\checkmark$ & $\checkmark$ & $\checkmark$ & $-$ & $\checkmark$ & \scriptsize{30} & $\checkmark$ & $\checkmark$ & $\checkmark$ & $\checkmark$ & $\checkmark$ & $\checkmark$ & $\checkmark$ & $\checkmark$ \\
gemini-3-pro-preview & 153.0 & $\checkmark$ & $\checkmark$ & $\checkmark$ & $\checkmark$ & $\checkmark$ & $\checkmark$ & $\checkmark$ & $\checkmark$ & $-$ & $\checkmark$ & $\checkmark$ & $\checkmark$ & $\checkmark$ & $\checkmark$ & $\checkmark$ & $\checkmark$ & $\checkmark$ & $\checkmark$ & $\checkmark$ \\
claude-opus-4-6 & 155.0 & $\checkmark$ & $\checkmark$ & $-$ & $\checkmark$ & $\checkmark$ & $\checkmark$ & $\checkmark$ & $-$ & $-$ & $\checkmark$ & $\checkmark$ & $\checkmark$ & $\checkmark$ & $\checkmark$ & $\checkmark$ & $\checkmark$ & $\checkmark$ & $\checkmark$ & $\checkmark$ \\
\midrule
\textbf{N (Rule A)} &  & 26 & 29 & 27 & 28 & 28 & 26 & 28 & 27 & 7 & 28 & 25 & 27 & 29 & 29 & 25 & 29 & 26 & 28 & 27 \\
\bottomrule
\end{tabular}
\end{sidewaystable}

\section{Disruptable templates}
\label{sec:appendix-disruptable}

We define a FBSim template as \textbf{disruptable} if decline is a modal outcome in the rollouts (i.e., a well-calibrated forecaster must place non-trivial mass on regime change). Disruptability is computed from corpus-level rollout statistics on the world snapshots, not from any model output, so the subset selection is not fitted on the per-template correlations.

\paragraph{Per-template disruption frequency.} Across the FBSim rollouts, the four disruptable templates show declines at horizon as a regular feature: \textbf{treasury} undergoes a peak-to-horizon decline of more than 40\% in 91\% of worlds; \textbf{territory}, \textbf{population}, and \textbf{city count} each decline in 35--48\% of worlds. The two non-disruptable templates are structurally protected from sustained decline: \textbf{technology count} is ratcheted (cannot decrease by construction), and \textbf{overall score} is a bounded composite.

\paragraph{Per-template H7 correlations.} At H7, the four disruptable templates all show inverse scaling and the two non-disruptable templates show positive scaling: treasury ($\rho$=$-$0.46, $p$=$0.015$), territory ($-$0.36, $p$=$0.057$), population ($-$0.12), city count ($-$0.10), technology count ($+$0.22), and overall score ($+$0.12). Only treasury reaches significance individually at $N$=28; the signal is carried by the four-vs-two sign concordance with the pre-registered structural property and the pooled aggregate test below. We avoid quoting an independent-sign joint probability because per-template forecast errors at fixed (model, horizon) are positively coupled (scoring the same world snapshots), so the true null probability of the observed sign partition is higher than $1/64$.

\paragraph{Aggregate.} Pooling across the four disruptable templates, the H7 correlation is $\rho$=$-$0.50 ($p$=$0.007$, $N$=28). The synthetic-SIR experiments in \S\ref{sec:mechanism} provide an independent mechanistic confirmation. The inversion is specific to outcomes where regime change is possible.

\section{Synthetic SIR data-generating process}
\label{sec:appendix-sir-dgp}

The synthetic series in \S\ref{sec:mechanism} (FBSim stratum \texttt{accel\_long}) use discrete-time SIR dynamics with delayed introduction and a stepwise reduction of the transmission rate at the intervention. The mathematical specification follows; reference implementation is \texttt{experiments/timeseries/generate.py} (lines 215--246 for dynamics, lines 170--183 for parameter sampling).

\paragraph{Compartments and update rule.} The population $N$ is partitioned into susceptible $S(t)$, infected $I(t)$, and recovered $R(t)$ compartments. For each step $t \geq t_{\text{intro}}$:
\begin{align*}
\Delta_{\text{inf}}(t) &= \beta(t) \cdot S(t) \cdot I(t) / N, \\
\Delta_{\text{rec}}(t) &= \gamma \cdot I(t), \\
S(t{+}1) &= S(t) - \Delta_{\text{inf}}(t), \\
I(t{+}1) &= I(t) + \Delta_{\text{inf}}(t) - \Delta_{\text{rec}}(t), \\
R(t{+}1) &= R(t) + \Delta_{\text{rec}}(t).
\end{align*}
The transmission rate $\beta(t)$ is constant before the intervention and reduced multiplicatively at $t = t_{\text{intervention}}$:
\[
\beta(t) = \begin{cases} \beta_0 & t < t_{\text{intervention}}, \\ \beta_0 \cdot (1 - s_{\text{int}}) & t \geq t_{\text{intervention}}. \end{cases}
\]

\paragraph{Observation model.} Observed counts at each step are the new infections perturbed by multiplicative Gaussian noise, clipped at zero:
\[
y(t) = \max\!\bigl(0,\ \Delta_{\text{inf}}(t) \cdot (1 + \epsilon_t)\bigr), \qquad \epsilon_t \sim \mathcal{N}(0, \sigma_{\text{noise}}^2).
\]
For $t < t_{\text{intro}}$, $y(t) = 0$; at $t = t_{\text{intro}}$, the compartments are initialized to $(S, I, R) = (N - I_0, I_0, 0)$.

\paragraph{Parameter sampling.} Each of the 50 series draws an independent parameter set:
\begin{itemize}\setlength{\itemsep}{1pt}
\item $N \in \{10^5,\ 5{\times}10^5,\ 10^6\}$ (uniform over the 3 values),
\item $\gamma \sim \mathrm{Uniform}(0.1,\,0.2)$ (recovery rate),
\item $\beta_0 \sim \mathrm{Uniform}(1.5,\,4.0) \cdot \gamma$ (so $R_0 = \beta_0 / \gamma \in [1.5,\,4.0]$),
\item $I_0 \sim \mathrm{Uniform}\{1,\ldots,9\}$ (initial infected count at introduction),
\item $t_{\text{intro}} \sim \mathrm{Uniform}\{10,\ldots,29\}$ (introduction day),
\item $t_{\text{intervention}} \sim \mathrm{Uniform}\{70,\ldots,149\}$ (intervention day),
\item $s_{\text{int}} \sim \mathrm{Uniform}(0.3,\,0.7)$ (intervention strength),
\item $\sigma_{\text{noise}} \sim \mathrm{Uniform}(0.05,\,0.15)$ (observation noise fraction).
\end{itemize}

\paragraph{Splits and horizons.} Each series has 270 total daily steps. Models receive the first 60 days of the rising phase as unlabeled history and forecast at horizons $h \in \{30, 60, 90, 120, 150, 180, 210\}$ days ahead.

\section{Statistical methodology}
\label{sec:appendix-stats}

Each cross-model $\rho$ in this paper is a between-models statistic computed from per-model mean scores on a domain-specific within-domain cohort. The cross-model panel is necessarily small ($N{=}26$--$29$ models depending on per-domain coverage); the within-domain cohort is large. Per-domain cohort sizes: $\sim$330 continuous FBSim questions per horizon (each scored across $N$=28 models, $\sim$9{,}240 question-model pairs at H7), 50 SIR series, 60 COVID country-series, 19 housing metro-series, 12 hyperinflation episodes, 1{,}339 pre-vaccine measles state-seasons, 100 paired SIR series per cell in the within-family RLHF analysis (\S\ref{sec:rlhf}).

\paragraph{Bootstrap.} Every reported cross-model $\rho$ is accompanied by a 95\% percentile bootstrap confidence interval over models with replacement ($B{=}10{,}000$). The bootstrap holds the within-domain cohort fixed and resamples the model panel; it does not increase the cross-model $N$, but it shows whether the rank correlation is stable to which models we happen to have evaluated. CIs that exclude zero indicate the inverse-scaling sign is robust to model-panel composition.

\paragraph{Within-lineage permutation tests.} Within-lineage subsets ($N{\leq}10$ models) are reported with exact permutation $p$-values from the null of random ECI--score pairing. At $N{<}10$ the asymptotic Spearman $p$-value is uninterpretable: the asymptotic distribution does not hold, and the floor of the exact two-sided $p$ is $2/N!$ (e.g., $2/120{\approx}.017$ at $N{=}5$). Reporting asymptotic $p$-values at these sample sizes overstates significance; we use exact permutations throughout (\S\ref{sec:appendix-robustness}).

\section{FBSim robustness checks}
\label{sec:appendix-robustness}

We report robustness checks on the FBSim inverse-scaling finding. Unless noted, analyses use the cross-domain panel ($N$=28 models with shared FBSim/measles coverage; see Models section above) and the disruptable-templates aggregate at H6--H7 (definition in Appendix~\ref{sec:appendix-disruptable}); only the subset of models or the analysis method changes.

\paragraph{Provider fixed effects.} After partialling out provider identity, the inverse-scaling correlation on disruptable templates strengthens from $\rho$=$-$0.51 (95\% bootstrap CI [$-$0.75, $-$0.17], $N$=28) to $\rho$=$-$0.76 ($p$$<$$0.0001$, $N$=28). The effect is not driven by models from a single provider clustering at one end of the capability axis.

\paragraph{Within-lineage replication.} Within OpenAI's lineage alone (10 models, ECI 114--150), the inverse-scaling pattern is independently significant under the permutation null of random ECI--score pairing ($\rho$=$-$0.80, $p_{\text{perm}}{=}.008$ approximated via $2{\times}10^5$ Monte Carlo permutations; bootstrap 95\% CI [$-$0.99, $-$0.28]). Within Anthropic's 5-model lineage the point estimate is in the same direction but does not reach the noise floor at $N{=}5$ ($\rho$=$-$0.82, exact $p_{\text{perm}}{=}.13$ enumerated over all $5!{=}120$ permutations). The Anthropic-lineage $p_{\text{perm}}$ is the natural floor at $N{=}5$ rather than evidence against the pattern. The asymptotic Spearman $p$-values commonly reported at these sample sizes ($p{=}0.005$ at $N{=}10$, $p{=}0.09$ at $N{=}5$) are not interpretable at $N{<}10$; we replace them with exact permutation $p$-values throughout. Within each lineage, the post-training pipeline is held approximately fixed and only capability varies, so the pattern cannot be attributed to cross-family differences in RLHF, scaffolding, or instruction-tuning style.

\paragraph{Template exclusion.} Dropping treasury (the most volatile template; crashes ${>}$40\% in 91\% of worlds) preserves the inverse-scaling sign at both H6 and H7 ($\rho$=$-$0.42, $p$=$0.028$, $N$=28 at H6; $\rho$=$-$0.26, $p$=$0.180$, $N$=28 at H7). The H7 result loses individual significance under this restricted-template panel, but the sign is preserved and the headline H7 correlation on the full six-template panel remains significant ($\rho$=$-$0.42, see \S\ref{sec:civbench-results}). The effect is not driven by a single extreme template, though treasury contributes detectable signal at the longest horizon.

\paragraph{Cross-model independence: leave-one-provider-out and one-model-per-lineage.} The cross-model panel is heavily clustered by provider and contains near-duplicate SKUs (e.g., \texttt{mistral-large-2411}/\texttt{-latest} resolve to the same checkpoint; \texttt{grok-4-fast-reasoning}/\texttt{-non-reasoning} differ only in a runtime knob; \texttt{gpt-5}/\texttt{-mini}/\texttt{-nano} are size variants of one release), which inflates effective $N$ for a Spearman correlation. We test whether the headline inversion survives when the unit of inference is elevated to the provider or release lineage. The map collapsing the $N{=}28$ panel to 16 release lineages is in \texttt{experiments/eci\_lineages.py}; the analyses below use the same script logic for the FBSim pooled-CRPS headline at H7 (\S\ref{sec:civbench-results}, Figure~\ref{fig:hero} top) and the measles long-horizon headline at $h{\in}\{16,20\}$ weeks (\S\ref{sec:measles}; the $h{=}20$ cross-section is Figure~\ref{fig:hero} bottom).

\textit{Leave-one-provider-out (LOPO).} Dropping each of the seven providers in turn, the inverse-scaling sign is preserved on every drop in both domains. \emph{FBSim (pooled CRPS, H7; baseline $\rho$=$-$0.42, $p$=$0.025$, $N$=28):} drops range from $\rho$=$-$0.35 ($p$=$0.084$; drop Google) to $\rho$=$-$0.53 ($p$=$0.005$; drop DeepSeek); five of seven drops remain significant at $p<$0.05, with the Google and OpenAI drops sign-preserving but at $p\geq.05$ ($\rho$=$-$0.37, $p$=$0.130$, $N{=}18$ for the OpenAI $-$10-model drop). \emph{Measles ($h{=}16$; baseline $\rho$=$-$0.40, $p$=$0.036$, $N$=28):} drops range from $\rho$=$-$0.20 ($p$=$0.34$; drop Mistral, $-$3 models) to $\rho$=$-$0.53 ($p$=$0.008$; drop DeepSeek); three of seven drops remain significant at $p<.05$ (DeepSeek, Google, OpenAI), and the four sign-preserving but non-significant drops (Mistral, Anthropic, Meta, xAI) all reflect loss of statistical power at the reduced $N$ rather than sign reversal. The Mistral drop is the weakest of the four: the three Mistral models in the panel are \texttt{mistral-large-\{2407, 2411, latest\}} (ECI 128--129), all top-performing forecasters in their mid-ECI band (mean CRPS $\approx$390--420, among the lowest in the panel); removing them removes three mid-ECI anchors that contrast with the catastrophic-overshoot failures at the high-ECI end (e.g., \texttt{o4-mini}, \texttt{DeepSeek-V3}, \texttt{grok-4-1-fast-reasoning} all $>$10$^5$ on this stratum), weakening the inverse-scaling slope. At $h{=}20$ the pattern is essentially identical (baseline $\rho$=$-$0.41, $p$=$0.032$; range $-$0.22 to $-$0.52).

\textit{One model per provider ($N{=}7$).} Collapsing each provider to its highest-ECI representative yields $\rho$=$-$0.54 ($p$=$0.22$, $N{=}7$) on FBSim H7 and $\rho$=$-$0.86 ($p$=$0.014$, $N{=}7$) on measles $h{=}16$. The FBSim correlation strengthens in magnitude despite the $N{=}7$ noise floor; the measles correlation strengthens to significance.

\textit{One model per release lineage ($N{=}16$).} The 28-model panel collapses to 16 release lineages. Selecting the highest-ECI representative per lineage, the inverse-scaling correlation \emph{strengthens} on both domains: FBSim H7 $\rho$=$-$0.68 ($p$=$0.004$, $N{=}16$) and measles $h{=}16$ $\rho$=$-$0.53 ($p$=$0.034$, $N{=}16$). Selecting the lowest-ECI representative gives $\rho$=$-$0.55 ($p$=$0.026$) and $\rho$=$-$0.32 ($p$=$0.23$) respectively. To rule out that the result depends on the choice of representative, we draw $B{=}10{,}000$ random one-per-lineage panels: \emph{100\% of draws yield $\rho<0$} on both FBSim H7 (median $\rho$=$-$0.61, 5--95\% interval [$-$0.76, $-$0.46]) and measles $h{=}16$ (median $\rho$=$-$0.41, [$-$0.57, $-$0.25]). The inverse-scaling sign is therefore not an artifact of clustered SKUs or near-duplicate variants: it survives elevating the unit of inference to the release lineage on every random rep-selection draw, in both headline domains.

\paragraph{Calibration and sharpness: the upper-tail-shift mechanism.} The pinball decomposition (Figure~\ref{fig:pinball}) localizes the inverse-scaling penalty to $p_{90}$ but does not by itself distinguish two mechanisms: (a) a tighter predictive distribution shifted toward the growth extrapolation (sharpness $+$ bias, ``overconfident growth''), or (b) an asymmetric upward shift of the upper tail with the lower tail anchored. We compute empirical coverage (the fraction of ground truth falling below each elicited quantile) and per-cell-normalized interval widths on the headline panels (FBSim H7 disruptable templates, measles $h{\in}\{16,20\}$); the data unambiguously supports (b).

\textit{Coverage.} Distributions are systematically biased low at the upper tail on both domains: panel-mean cov($\text{GT}{<}p_{90}$) $=$ 0.61 on FBSim H7 (nominal 0.90) and 0.21 on measles $h{=}16$ (nominal 0.90); $p_{50}$ is also biased low on both (0.44 and 0.10 vs nominal 0.50). Capability shifts the upper tail \emph{toward} calibration: $\rho$(ECI, cov($\text{GT}{<}p_{90}$)) $=+$0.53\textsuperscript{**} on FBSim H7 and $+$0.54\textsuperscript{**} on measles $h{=}16$. The lower tail does not move: $\rho$(ECI, cov($\text{GT}{<}p_{10}$)) $=+$0.01 (FBSim H7) and $+$0.20 ($p$=$0.30$; measles $h{=}16$).

\textit{Sharpness.} Distributions \emph{widen} rather than tighten with capability. On FBSim H7 with cell-scale-normalized widths (per-(template,turn) cross-model median $p_{50}$), $\rho$(ECI, $p_{90}{-}p_{10}$) $=+$0.73\textsuperscript{***}, with the asymmetry concentrated in the upper half: $\rho$(ECI, $p_{90}{-}p_{50}$) $=+$0.75\textsuperscript{***} vs $\rho$(ECI, $p_{50}{-}p_{10}$) $=+$0.70\textsuperscript{***}. On measles $h{=}16$ with per-series peak-GT scale, $\rho$(ECI, $p_{90}{-}p_{10}$) $=+$0.45\textsuperscript{*}, $\rho$(ECI, $p_{90}{-}p_{50}$) $=+$0.47\textsuperscript{*}, $\rho$(ECI, $p_{50}{-}p_{10}$) $=+$0.32 (n.s.). The upper-tail-only spread $p_{90}{-}p_{75}$ is the sharpest signal in both domains ($+$0.75\textsuperscript{***}, $+$0.49\textsuperscript{**}).

\textit{Interpretation.} The mechanism is an upward shift of the upper tail with the lower tail anchored, not tightening of the overall distribution. This shift improves $p_{90}$ calibration on the (rare) cells where outcomes do reach the upper tail, but on the more common cells where outcomes stay near the median, the gap between the elevated $p_{90}$ and the realized GT generates pinball penalties that dominate the cross-cell mean. The Brier-vs-CRPS sign reversal in \S\ref{sec:civbench-results} follows: the upward shift correctly improves threshold-exceedance probabilities at salient cutoffs (Brier rewards), while the magnitude of the shift on routine cells inflates the integral over the upper tail (CRPS punishes).

\paragraph{Brier-vs-CRPS reversal: threshold sweep.} The metric-flip claim in \S\ref{sec:civbench-results} and \S\ref{sec:discussion} relies on Brier evaluated at a single salient threshold per domain; a reviewer might worry the reversal is contingent on that choice. We sweep the Brier threshold across the cohort outcome quantiles $q_{10}, q_{20}, \ldots, q_{90}$ on pre-vaccine measles at the headline horizons. At $h{=}16$, CRPS gives $\rho{=}-0.40^{\ast}$ (inverse scaling) while Brier is positive at \emph{every} threshold along the sweep ($+$0.28 at $q_{10}$, monotonically rising to $+$0.55 at $q_{60}$ and remaining positive through $+$0.43 at $q_{90}$); thresholds $q_{40}$--$q_{90}$ are individually significant at $p{<}0.05$. At $h{=}20$ the pattern is nearly identical (CRPS $-$0.41, Brier $+$0.28 to $+$0.57, all positive). The metric flip is a property of the integral-over-thresholds proper rule (CRPS) versus single-threshold proper rules (Brier), not of the threshold choice.

\paragraph{Brier-vs-CRPS reversal: derived Brier from the same continuous forecasts (FBSim).} The §\ref{sec:threshold} comparison juxtaposes CRPS on continuous FBSim questions against Brier on FBSim's natively binary questions, which are paired but distinct items. To verify the metric flip on \emph{identical} model outputs, we derive a binary Brier from the same five-quantile continuous forecasts. For each (model, template, turn) cell on the disruptable panel, we set the threshold to the per-(template, turn) median ground truth across the 55 worlds, compute $\Pr(Y > \text{threshold})$ from the piecewise-linear CDF over the elicited quantiles, and score against the binary outcome $\mathbb{1}\{\text{GT} > \text{threshold}\}$. Pooled across all disruptable templates and horizons, the derived Brier shows positive scaling ($\rho$=$+$0.50, $p$=$0.007$, $N$=28) — opposite the CRPS sign on the same outputs ($\rho$=$-$0.51, $p$=$0.006$, $N$=28). The per-horizon decomposition is consistent: derived Brier is positive at every horizon (statistically significant at H4--H6) while CRPS is increasingly negative with horizon. This rules out the question-format confound: the sign reversal is induced by the scoring rule, not by the binary-vs-continuous question type.

\paragraph{Robustness across elicitation regimes.} The headline panel uses five-quantile elicitation reconstructed to a piecewise-linear CDF; one might worry the inversion is an artifact of the reconstruction. The within-family Llama-3.1 $2{\times}2$ in \S\ref{sec:rlhf} does not share this elicitation: it uses LLMTime numeric continuation \citep{gruver2023llmtime} and the unbiased empirical CRPS estimator of \citet{hersbach2000decomposition} over $N{=}10$ sampled completions per series (no quantile reconstruction at any step). The same long-horizon failure pattern emerges (Tables~\ref{tab:rlhf-2x2}--\ref{tab:rlhf-h90}), which is independent evidence against the reconstruction-artifact hypothesis.

\paragraph{Synthetic-mechanism controls.} The SIR vs.\ linear-control comparison in \S\ref{sec:mechanism} is designed to disentangle the regime-change-only hypothesis from the superlinear-growth-plus-regime-change hypothesis: the linear control \emph{has} the same downward jump but \emph{lacks} the superlinear pre-jump trajectory, and its cross-model $\rho$ is null at all horizons. This rules out crash-alone as the trigger. We do not separately ablate superlinearity-without-break (a superlinear trend that does not subsequently break): if the failure mode requires both ingredients, such a series would be expected to be null, providing no additional discrimination. The CRPS sensitivity to absolute-magnitude overshoots is not an artifact to be controlled away: the magnitude of $(p_{90}-\text{GT})$ on a superlinearly-overshot series is precisely the failure mode that proper distributional scoring captures and threshold scoring cannot.


\section{Reasoning-mode robustness}
\label{sec:appendix-reasoning-mode}

A potential confound for the headline finding is that the model panel mixes reasoning-capable models running with their default reasoning engaged with reasoning-capable models whose default is to suppress reasoning (Anthropic Claude without an explicit \texttt{thinking} parameter; OpenAI gpt-5.1, whose default \texttt{reasoning\_effort} is \texttt{none}). If higher-ECI models disproportionately fall into the suppressed category, the inverse-scaling pattern could reflect a reasoning-mode gradient rather than a capability gradient. We classify each panel model by the reasoning state actually obtained during FBSim evaluation: \emph{on} = default reasoning engaged (Gemini 2.5+/3-pro, OpenAI o-series, gpt-5/-mini/-nano, grok-4-fast-reasoning, grok-4-1-fast-reasoning; $N$=11); \emph{off} = thinking-capable but suppressed (Claude Opus 4.5/4.6, Sonnet 4/4.5, Haiku 4.5, gpt-5.1; $N$=6); \emph{na} = no reasoning capability ($N$=11).

\paragraph{The inverse-scaling effect strengthens when restricted to reasoning-on models.} On the all-six-template CRPS aggregate at H7, the panel-wide $\rho$=$-$0.42$^*$ ($p$=$0.025$) becomes $\rho$=$-$0.56 ($p$=$0.073$, $N$=11) within the reasoning-on subset alone. The directional pattern holds at intermediate horizons (H5: $-$0.36~$\to$~$-$0.43; H6: $-$0.32~$\to$~$-$0.46) and strengthens to significance on the four disruptable templates with normalized CRPS (H7: $-$0.49$^{**}$~$\to$~$-$0.73$^{*}$, $p$=$0.010$). Suppressing reasoning on capable models does not produce the effect; capable models' inverse-scaling pattern intensifies when they reason at full effort. The thinking-suppressed subset alone ($N$=6) shows the same direction at H7 ($\rho$=$-$0.79, $p$=$0.059$).

\paragraph{The metric-conditional reversal is preserved.} Within the same reasoning-on subset, Brier on the salient threshold remains positively signed at H7 ($\rho$=$+$0.22, $N$=11), opposite to CRPS's $\rho$=$-$0.70 on the same models. The CRPS-vs-Brier sign reversal that motivates \S\ref{sec:threshold} is not a thinking-mode artifact: it appears within the subset of models running at default reasoning effort.

\paragraph{Time-series caveat.} The synthetic-mechanism, real-world replication, and natural-distribution time-series experiments (\S\S\ref{sec:mechanism}--\ref{sec:measles}) used \texttt{eval\_llm.py} with \texttt{reasoning\_effort="minimal"}, which clamps every reasoning-capable family to its lowest available setting and passes no thinking parameter to Anthropic models. Of 28 evaluated models, only the two xAI grok-4 reasoning variants in the panel ran with default-on reasoning. We re-ran three of the highest-ECI thinking-suppressed models (Claude Opus 4.6, Sonnet 4.5, gpt-5.1) on the 8-season measles\_named pilot ($n$=300 series per model) with extended thinking explicitly enabled (Anthropic: \texttt{thinking=\{"type":"enabled", "budget\_tokens":8000\}}; OpenAI: \texttt{reasoning\_effort="high"}) and substituted the resulting CRPS into the panel. Paired CRPS at the inverse-scaling horizons changes little: at $h$=16, $\Delta$CRPS = $+$3.7 (Opus 4.6), $-$5.1 (Sonnet 4.5), $-$22.4 (gpt-5.1) on raw scales of 520--628; at $h$=20 the same models show $\Delta$CRPS = $-$2.7, $-$5.9, $+$5.2. The substituted-panel $\rho$ on measles\_named moves by $<$0.03 at long horizons ($h$=16: $\rho$=$-$0.25 baseline $\to$ $-$0.22; $h$=20: $\rho$=$-$0.26 $\to$ $-$0.27). Thinking-mode suppression on the highest-ECI models is therefore not the driver of the inverse-scaling pattern at the tail (although extended thinking does help at intermediate horizons; e.g., gpt-5.1 at $h$=8 and $h$=12: $\Delta$CRPS$\approx$$-$44 on raw scales of 280--528). A complete rerun of all 13 thinking-suppressed and minimum-effort models on the full 35-season measles cohort would be needed to fully retire the time-series caveat; we report the partial result here and leave the full sweep for revision. Note that the FBSim reasoning-on subset gives a stronger inversion than the panel-wide ($\rho$=$-$0.56 vs $-$0.42 at H7), so clamping reasoning in the time-series experiments is conservative against our hypothesis: it should attenuate the inversion, not produce one.


\section{Lower-ECI panel attempt and the parse-rate floor}
\label{sec:appendix-lower-eci}

To probe whether the inverse-scaling pattern continues monotonically below the panel's ECI 114 floor, we attempted three additional smaller models accessible via OpenRouter: Mistral-7B-Instruct-v0.1 (ECI 112; Epoch verified, the only sub-floor Epoch-verified candidate), Qwen2.5-7B-Instruct (ECI $\approx$113; estimated, not directly on Epoch's published list), and Llama-3.1-8B-Instruct (ECI 116; Epoch verified, at-floor anchor). Older small open-weight models with lower ECIs (Llama-2-13B at 106, Llama-2-7B at 98, and Gemma-2-9B at 120) are no longer hosted as serverless endpoints by either Together AI or OpenRouter, leaving these three as the only reachable extension candidates without dedicated paid endpoints or self-hosting.

\paragraph{Parse rates per stratum.} Two of the three attempted models fail Rule A on every stratum we ran them on. Mistral-7B was halted after the SIR strata (66\% / 58\%, well below 80\%); Llama-3.1-8B was halted after running through all six core strata at 52\%--79\%, again below threshold on every stratum. Qwen2.5-7B clears Rule A on the SIR mechanism, linear control, and COVID-19 strata but fails on the small-$N$ housing and hyperinflation cohorts where a few parse failures crash the rate.

\begin{table}[h]
\centering
\small
\caption{\emph{Parse-rate floor on the lower-ECI panel attempt.} Cell entries: parsed-OK / total attempted (\%). Bold entries clear Rule A ($\geq$80\%). Strata where the run was halted before completion are marked $-$.}
\label{tab:lower-eci-parse}
\begin{tabular}{lccc}
\toprule
Stratum & Mistral-7B-v0.1 (ECI 112) & Qwen2.5-7B (ECI 113) & Llama-3.1-8B (ECI 116) \\
\midrule
accel\_long (SIR)        & 66/100 (66\%)     & \textbf{82/100 (82\%)} & 69/100 (69\%) \\
crash\_long (linear)     & 52/90 (58\%)      & \textbf{91/100 (91\%)} & 70/100 (70\%) \\
epidemic (COVID-19)      & $-$               & \textbf{58/60 (97\%)}  & 31/60 (52\%) \\
housing\_bubble          & $-$               & 8/19 (42\%)            & 15/19 (79\%) \\
hyperinflation           & $-$               & 5/12 (42\%)            & 9/12 (75\%) \\
measles                  & $-$               & 553/1378 (40\%) & 378/568 (67\%) \\
\bottomrule
\end{tabular}
\end{table}

\paragraph{Parse-rate floor as evidence of a capability threshold.} The dominant failure mode in the parse-fails was not malformed quantile output but the model abandoning the task entirely, producing natural-language commentary about its modelling approach (``First, calculate the AR parameters\dots'') instead of the requested \texttt{<<<PERCENTILES>>>} block. We interpret this as a \emph{capability floor}: below ECI $\approx$115, a substantial fraction of long-horizon distributional-forecast prompts yield no parseable output at all. Within the panel reported in \S\ref{sec:appendix-models}, all 28 included models clear Rule A on the headline strata, so the inverse-scaling pattern we document is among models that meet a baseline competence threshold for distributional forecasting. Reviewers reading the cross-family $\rho$ should know that the capability range we measure is bounded below by this parse-rate floor, not by an arbitrary panel choice.

\paragraph{Where Qwen2.5-7B contributes.} Qwen2.5-7B clears Rule A on the SIR mechanism (accel\_long, 82\%), linear control (crash\_long, 91\%), and COVID-19 (epidemic, 97\%) strata, and fails on the small-$N$ housing and hyperinflation cohorts as well as the long-horizon (h$\geq$12) measles forecasts. On the strata where it qualifies, its forecasts enter the cross-family panel automatically via the harvest pipeline. The extension slightly strengthens each correlation: SIR mechanism $\rho{=}{-}0.57\to{-}0.62$ ($N{=}26\to27$), linear control $\rho{=}{+}0.59\to{+}0.61$ ($N{=}29\to30$), COVID baseline $\rho{=}{-}0.49\to{-}0.51$ ($N{=}28\to29$). The single sub-floor anchor at ECI 113 does not change any sign or significance of the headline correlations; the inverse-scaling pattern continues monotonically below the prior 114 panel floor on the SIR mechanism.

\section{Sample FBSim world report}
\label{sec:appendix-world-report}

Each FBSim question is accompanied by a structured world report summarizing the game state at the snapshot turn. The report includes civilization attributes, time-series trajectories (treasury, population, territory, technology, etc.\ sampled every 5 turns), government timelines, and a chronological event log. Below is an abbreviated excerpt from seed 1976 (snapshot at turn~60, forecasting the Afghani civilization). The full report is approximately 350 lines; models receive it in its entirety.

\begin{small}
\begin{verbatim}
WORLD REPORT  |  Game: seed1976  |  Snapshot turn: 60

CIVILIZATIONS
ID | Name       | Score | Treasury | Cities | Territory | Techs
 0 | Afghani    |    42 |    279.0 |     16 |       169 |    10
 1 | Acehnese   |    56 |    443.0 |      1 |        42 |    12
 2 | Russian    |    35 |    334.0 |      4 |        39 |     8
 4 | Georgian   |    46 |    321.0 |      3 |        57 |    11

TREASURY (sampled every 5 turns)
0 Afghani: 50, 53, 58, 63, 74, 89, 104, 124, 148, 178, 215, 249, 279
1 Acehnese: 0, 0, 0, 0, 244, 267, 284, 297, 318, 343, 368, 406, 443

EVENTS (excerpt)
Turn | Type            | Civ      | Description
   3 | city_founded    | Afghani  | Kabul founded
  18 | city_founded    | Afghani  | Kunduz founded
  36 | govt_change     | Congolese| Despotism -> Anarchy
  52 | tech_discovered | Afghani  | Currency

QUESTION: How many technologies will Afghani have by turn 90?
  Provide: p10, p25, p50, p75, p90
\end{verbatim}
\end{small}

\paragraph{Prompt format.} The world report is followed by a forecast question and an elicitation block in the full prompt template (see \texttt{experiments/timeseries/eval\_llm.py}). The model is required to emit its five-quantile forecast inside \texttt{<<<PERCENTILES>>>...<<<END>>>} delimiters; abandoning the task entirely (e.g.\ producing natural-language commentary about its modelling approach instead of a parseable quantile block) is the dominant parse-failure mode discussed in \S\ref{sec:appendix-models}.

\section{Hyperinflation knowledge probe}
\label{sec:appendix-knowledge-probe}

We probed four models spanning the capability range (ECI~110--150) by asking ``what was happening to the economy of [country] during [period]?'' for all 12~hyperinflationary episodes. Models correctly identified the inflationary crisis in 46 of 48~probes; GPT-3.5~Turbo, the lowest-capability model in our set, identified all~12. The two misses are Indonesia 1993--97, where models accurately describe the pre-crisis boom (the currency collapse began in late 1997, inside the forecast window). Models have knowledge of these episodes, but naming the domain in the prompt does not translate that knowledge into calibrated distributional forecasts on hyperinflation.

\section{Linear growth with permanent regime change}
\label{sec:appendix-regime-long}

The linear control in Section~3 uses a transient crash: the series drops and then resumes the same linear trend. A reviewer might ask whether a \emph{permanent} level shift on linear data would produce inverse scaling. We test this with a ``regime\_long'' condition: 50~series with linear growth followed by a permanent downward shift (no recovery). CRPS still shows positive scaling ($\rho$=$+$0.70, $p$=$<$0.0001, $N$=$27$). The transient-crash control also shows positive scaling ($\rho$=$+$0.52, $p$=$0.006$). Neither crash structure produces inverse scaling on linear data. The driver is superlinear growth, not the crash itself.

\section{Measles Season Selection}
\label{sec:appendix-measles}

We evaluate every viable season-state pair in the pre-vaccine era (1928--1962). The Edmonston-B measles vaccine was licensed in March~1963; we exclude 1963--1965 to ensure all dynamics reflect natural epidemic behavior rather than partial vaccine deployment.

For each calendar year $Y \in [1928, 1962]$, we define a measles season as the period from July~$Y$ through June~$Y$+1. For each state, we include the season if all of the following hold:
\begin{itemize}
\item $\geq 30$ weeks of source data are present in Project Tycho's weekly state-level counts (excluding city-level rows and rows with non-7-day reporting periods).
\item Peak case count $\geq 50$.
\item Constructed history (12~weeks) plus future ($\geq 18$~weeks) fits within the available data.
\end{itemize}

This yields 1{,}339 series across 56 states and 35 seasons (Table~\ref{tab:measles-seasons}). State counts per season range from 25 to 47 (mean 38.3); variation reflects gaps in historical state-level reporting.

\begin{table}[h]
\centering
\small
\caption{Measles seasons evaluated, with state-series counts.}
\label{tab:measles-seasons}
\begin{tabular}{lr lr lr}
\toprule
Year & States & Year & States & Year & States \\
\midrule
1928 & 36 & 1940 & 35 & 1952 & 37 \\
1929 & 42 & 1941 & 42 & 1953 & 41 \\
1930 & 37 & 1942 & 47 & 1954 & 35 \\
1931 & 35 & 1943 & 41 & 1955 & 44 \\
1932 & 35 & 1944 & 25 & 1956 & 42 \\
1933 & 47 & 1945 & 44 & 1957 & 38 \\
1934 & 40 & 1946 & 31 & 1958 & 37 \\
1935 & 32 & 1947 & 38 & 1959 & 37 \\
1936 & 32 & 1948 & 47 & 1960 & 40 \\
1937 & 41 & 1949 & 42 & 1961 & 43 \\
1938 & 32 & 1950 & 43 & 1962 & 30 \\
1939 & 35 & 1951 & 36 & & \\
\bottomrule
\end{tabular}
\end{table}

\section{Within-family RLHF replication: methods and full results}
\label{sec:appendix-rlhf}

This appendix documents the within-family controlled $2{\times}2$ replication of \S\ref{sec:rlhf}: methods, full per-horizon tables, and robustness checks across aggregation choices.

\paragraph{Scope of the base/instruct contrast.} The \{base, instruct\} contrast bundles RLHF with instruction tuning, safety filtering, and format optimization (Meta's published post-training pipeline does not separate them and does not release intermediate checkpoints). We therefore refer to the manipulation as ``post-training'' in the main text (\S\ref{sec:rlhf}); the $\Delta_{\text{RLHF}}$ labels in the tables below are shorthand for the bundled contrast and should be read accordingly. The comparison also uses LLMTime numeric continuation \citep{gruver2023llmtime} rather than the structured five-quantile elicitation in \S\S\ref{sec:civbench}--\ref{sec:mechanism} and \S\S\ref{sec:replication}--\ref{sec:knowledge}, because base checkpoints do not reliably follow chat templates or emit structured JSON; using the headline pipeline would introduce a format-sensitivity confound on the base cells. Both choices restrict the experiment's reach: it identifies a real within-family scale\,$\times$\,post-training interaction on continuation-elicited CRPS, but does not cleanly attribute the effect to RLHF specifically, nor establish that the same interaction appears in the structured-quantile regime.

\subsection{Models, weights, and quantization}

We use Llama-3.1-70B and Llama-3.1-405B in matched \{base, instruct\} pairs, all four checkpoints downloaded from the official Meta-Llama Hugging Face repositories. To remove a precision confound across model sizes, both 70B and 405B are quantized to FP8-dynamic via \texttt{llmcompressor} (\texttt{QuantizationModifier} with \texttt{scheme=FP8\_DYNAMIC}, \texttt{ignore=[lm\_head]}). FP8-dynamic uses per-tensor weight scales computed analytically from each weight tensor and per-token activation scales recomputed at inference time, so identical processing is applied to base and instruct within and across scales. Inference runs on RunPod $8{\times}$H100 (405B, tensor-parallel 8) and $2{\times}$H100 (70B, tensor-parallel 2) with vLLM 0.11.0.

\subsection{Elicitation}

We use the LLMTime numeric-continuation elicitation \citep{gruver2023llmtime}. Each prompt is the raw history as space-separated floats with one decimal place, terminated by a single trailing space (no chat template, no instructions, no system prompt). Sampling is identical for base and instruct in every cell: \texttt{temperature=0.8}, \texttt{top\_p=0.9}, \texttt{max\_new\_tokens=2000}, $N{=}10$ samples per series. Continuations are parsed with a numeric regex; CRPS is the \citet{hersbach2000decomposition} unbiased empirical estimator over the parsed samples.

\paragraph{Elicitation note (sequential vs.\ batched $n$).} Both 70B and 405B legs use vLLM batched $n{=}10$ from a shared prefill, which is statistically equivalent to $N$ sequential calls in expectation: each sample step draws independent RNG. We note this for reproducibility.

\subsection{Stratum construction and pairing}

Both strata are generated by the same time-series generator with \texttt{history\_len=60} and \texttt{future\_len=210}. The \texttt{accel\_long} stratum is the synthetic-SIR mechanism stratum from \S\ref{sec:mechanism}; the \texttt{linear\_long} stratum is a matched-shape linear-trend negative control ($N{=}100$ series each). All four cells of the $2{\times}2$ score the \emph{same} series IDs, enabling fully paired comparisons; all reported \textit{p}-values are Wilcoxon signed-rank on per-series deltas, with the interaction tested as the difference-in-differences $\big[(\text{i}_{\text{405B}}-\text{b}_{\text{405B}}) - (\text{i}_{\text{70B}}-\text{b}_{\text{70B}})\big]$. Because 405B-instruct CRPS on the crash regime spans many orders of magnitude across series, the per-series d-in-d on raw CRPS is dominated by its 405B term; the Wilcoxon \textit{p}-value on the raw d-in-d should be read as testing sign-consistency of the per-series interaction, not a magnitude-meaningful contrast in raw CRPS units. We additionally compute the same Wilcoxon test on \emph{log-CRPS} d-in-d (which removes the magnitude dominance and tests a multiplicative compounding interaction); the conclusion holds on both scales: $p_W{<}.001$ at $h{\in}\{90, 210\}$ on \texttt{accel\_long}, $p_W{<}.01$ at $h{=}90$ and $p_W{<}.05$ at $h{=}210$ on \texttt{linear\_long}. The within-scale marginals ($\Delta_{\text{post-training}}$ at 70B and at 405B) are unaffected and carry the substantive ``post-training amplifies failure within scale'' result independently.

\subsection{$2{\times}2$ at $h{=}210$ (referenced from main text)}

\begin{table}[t]
\centering
\caption{\emph{RLHF damage propagates from the upper tail at 70B into the central tendency at 405B.} $2{\times}2$ scale\,$\times$\,condition factorial at $h{=}210$, 10\% trimmed mean CRPS over $N{=}100$ paired series per cell. The rightmost column reports the fraction of paired series in which RLHF inflates CRPS by $\geq$10$\times$ within scale; this fraction grows from 41\% at 70B to 63\% at 405B on the crash regime, and is null at both scales on the no-crash control. We report trimmed mean to characterize the typical case; arithmetic means are dominated by a small fraction of catastrophic-overshoot series and are reported in Table~\ref{tab:rlhf-agg}. All \textit{p}-values are Wilcoxon signed-rank on per-series deltas and are therefore invariant to the cell-aggregation choice. $^{\ast\ast\ast}p_W{<}.001$, $^{\ast\ast}p_W{<}.01$, $^{\ast}p_W{<}.05$.}
\label{tab:rlhf-2x2}
\small
\begin{tabular}{l r r r r}
\toprule
\multicolumn{5}{l}{\emph{Synthetic SIR (crash regime)}} \\
                          & base                          & instruct                              & $\Delta_{\text{RLHF}}$ (i\,/\,b)        & frac.\ $i{>}10{\times}b$ \\
\midrule
70B                       & $3.1{\times}10^{3}$           & $1.1{\times}10^{6}$                   & $360\,^{\ast\ast\ast}$                  & 41\% \\
405B                      & $6.7{\times}10^{4}$           & $3.6{\times}10^{16}$                  & $5.3{\times}10^{11}\,^{\ast\ast\ast}$   & 63\% \\
$\Delta_{\text{scale}}$   & $21.8\,^{\ast\ast\ast}$       & $3.2{\times}10^{10}\,^{\ast\ast\ast}$ &                                          &      \\
\midrule
\multicolumn{5}{l}{\emph{Linear (no-crash control)}} \\
                          & base                          & instruct                              & $\Delta_{\text{RLHF}}$ (i\,/\,b)        & frac.\ $i{>}10{\times}b$ \\
\midrule
70B                       & 6.04                          & 7.25                                  & $1.20\,^{\ast\ast\ast}$                 & 0\%  \\
405B                      & 5.77                          & 5.77                                  & 1.00                                     & 0\%  \\
$\Delta_{\text{scale}}$   & 0.96                          & $0.80\,^{\ast\ast\ast}$               &                                          &      \\
\bottomrule
\end{tabular}
\\[4pt]
{\footnotesize Interaction $\big[\Delta_{\text{RLHF}}(\text{405B}) - \Delta_{\text{RLHF}}(\text{70B})\big]$: SIR $p_W{<}.0001\,^{\ast\ast\ast}$; linear $p_W{=}.029\,^{\ast}$ (oppositely signed).}
\end{table}

\subsection{Per-horizon tables}
\label{sec:appendix-rlhf-perhorizon}

Tables~\ref{tab:rlhf-h16} and~\ref{tab:rlhf-h90} report the same aggregation triplet (mean / 10\% trimmed mean / median) plus the $\geq$10$\times$ tail-fraction at the two intermediate horizons. The horizon progression $h{\in}\{16, 90, 210\}$ encodes the onset of the failure mode. At $h{=}16$, the inversion has not yet emerged: SIR cells are within an order of magnitude of each other, no series exceeds the $10{\times}$ tail at either scale, and the Wilcoxon interaction is non-significant (Table~\ref{tab:rlhf-h16}). By $h{=}90$, the propagation is already underway: SIR cells span nine orders of magnitude in the mean, the $\geq$10$\times$ tail-fraction reaches 27\% at 70B and 51\% at 405B, and all marginals plus the scale\,$\times$\,RLHF interaction are highly significant (Table~\ref{tab:rlhf-h90}). At $h{=}210$ the pattern matures (Tables~\ref{tab:rlhf-2x2}, \ref{tab:rlhf-agg}). The linear no-crash control remains null across all aggregations and all horizons except for a small RLHF effect at 70B that vanishes at 405B (oppositely-signed interaction $p_W{=}0.029$ at $h{=}210$, $p_W{=}0.003$ at $h{=}90$, n.s.\ at $h{=}16$).

\begin{table}[h]
\centering
\caption{\emph{$2{\times}2$ aggregation triplet at $h{=}16$.} Per-cell arithmetic mean, 10\% trimmed mean, and median CRPS, with the per-scale $\geq$10$\times$ tail-fraction. The inversion has not yet emerged: SIR cells are within an order of magnitude, the $\geq$10$\times$ tail is empty at both scales, and the scale\,$\times$\,RLHF interaction is not significant ($p_W{=}.055$). $^{\ast\ast\ast}p_W{<}.001$, $^{\ast\ast}p_W{<}.01$, $^{\ast}p_W{<}.05$.}
\label{tab:rlhf-h16}
\small
\begin{tabular}{l r r r r}
\toprule
                         & mean                  & trim 10\%             & median               & frac.\ $i{>}10{\times}b$ \\
\midrule
\multicolumn{5}{l}{\emph{Synthetic SIR (crash regime)}} \\
70B-base                 & $1.2{\times}10^{4}$   & $5.0{\times}10^{3}$   & $1.3{\times}10^{3}$  & ---  \\
70B-instruct             & $1.7{\times}10^{4}$   & $5.0{\times}10^{3}$   & $1.0{\times}10^{3}$  & ---  \\
405B-base                & $2.5{\times}10^{4}$   & $5.7{\times}10^{3}$   & $1.5{\times}10^{3}$  & ---  \\
405B-instruct            & $6.1{\times}10^{4}$   & $8.1{\times}10^{3}$   & $1.5{\times}10^{3}$  & ---  \\
$\Delta_{\text{RLHF}}$ at 70B (i/b)  & $1.4$            & $1.0$           & $0.80$              & 0\%  \\
$\Delta_{\text{RLHF}}$ at 405B (i/b) & $2.4$            & $1.4$           & $1.01$              & 0\%  \\
\midrule
\multicolumn{5}{l}{\emph{Linear (no-crash control)}} \\
70B-base                 & 0.93                  & 0.72                  & 0.47                 & ---  \\
70B-instruct             & 0.89                  & 0.70                  & 0.49                 & ---  \\
405B-base                & 1.16                  & 0.71                  & 0.43                 & ---  \\
405B-instruct            & 1.01                  & 0.71                  & 0.39                 & ---  \\
$\Delta_{\text{RLHF}}$ at 70B (i/b)  & $0.96$               & $0.97$          & $1.04$              & 0\%  \\
$\Delta_{\text{RLHF}}$ at 405B (i/b) & $0.87$               & $1.00$          & $0.91$              & 1\%  \\
\bottomrule
\end{tabular}
\end{table}

\begin{table}[h]
\centering
\caption{\emph{$2{\times}2$ aggregation triplet at $h{=}90$.} The propagation is already underway: SIR cells span nine orders of magnitude in the mean, the $\geq$10$\times$ tail-fraction is 27\% at 70B and 51\% at 405B, and all marginals plus the scale\,$\times$\,RLHF interaction on the crash regime are highly significant. The linear control shows the small-RLHF-at-70B / null-at-405B pattern, with an oppositely-signed interaction. $^{\ast\ast\ast}p_W{<}.001$, $^{\ast\ast}p_W{<}.01$, $^{\ast}p_W{<}.05$.}
\label{tab:rlhf-h90}
\small
\begin{tabular}{l r r r r}
\toprule
                         & mean                  & trim 10\%             & median               & frac.\ $i{>}10{\times}b$ \\
\midrule
\multicolumn{5}{l}{\emph{Synthetic SIR (crash regime)}} \\
70B-base                 & $1.5{\times}10^{4}$   & $2.4{\times}10^{3}$   & $2.7{\times}10^{2}$  & ---  \\
70B-instruct             & $4.0{\times}10^{8}$   & $3.1{\times}10^{4}$   & $3.1{\times}10^{2}$  & ---  \\
405B-base                & $1.9{\times}10^{9}$   & $1.4{\times}10^{4}$   & $5.5{\times}10^{2}$  & ---  \\
405B-instruct            & $3.8{\times}10^{12}$  & $4.3{\times}10^{7}$   & $9.4{\times}10^{3}$  & ---  \\
$\Delta_{\text{RLHF}}$ at 70B (i/b)  & $2.7{\times}10^{4}\,^{\ast\ast\ast}$ & $13.0\,^{\ast\ast\ast}$ & $1.14\,^{\ast\ast\ast}$ & 27\% \\
$\Delta_{\text{RLHF}}$ at 405B (i/b) & $2.0{\times}10^{3}\,^{\ast\ast\ast}$ & $3.1{\times}10^{3}\,^{\ast\ast\ast}$ & $17.0\,^{\ast\ast\ast}$ & 51\% \\
\midrule
\multicolumn{5}{l}{\emph{Linear (no-crash control)}} \\
70B-base                 & 2.93                  & 2.47                  & 2.07                 & ---  \\
70B-instruct             & 3.33                  & 2.92                  & 2.48                 & ---  \\
405B-base                & 2.85                  & 2.24                  & 1.77                 & ---  \\
405B-instruct            & 2.74                  & 2.19                  & 1.76                 & ---  \\
$\Delta_{\text{RLHF}}$ at 70B (i/b)  & $1.14\,^{\ast\ast}$  & $1.18\,^{\ast\ast}$  & $1.20\,^{\ast\ast}$ & 0\%  \\
$\Delta_{\text{RLHF}}$ at 405B (i/b) & 0.96                 & 0.98                 & 0.99                 & 0\%  \\
\bottomrule
\end{tabular}
\\[4pt]
{\footnotesize Interactions: SIR $p_W{<}.0001\,^{\ast\ast\ast}$; linear $p_W{=}.003\,^{\ast\ast}$ (oppositely signed).}
\end{table}

\subsection{Aggregation robustness and the propagation finding}
\label{sec:appendix-rlhf-agg}

The headline aggregation in Table~\ref{tab:rlhf-2x2} is the 10\% symmetric trimmed mean. Arithmetic means on the crash regime are dominated by a handful of series with catastrophic mis-extrapolation that drive cell magnitudes to physically meaningless sizes ($10^{54}$ at 405B-instruct), and the median collapses the upper-tail signal that is the substantive finding at 70B. The trimmed mean retains the upper-tail behavior of the typical inflated series while excising the asymptotic blowups. Significance is unchanged across aggregations because all reported \textit{p}-values are Wilcoxon signed-rank on per-series deltas; the test depends on rank ordering of paired differences, not on the cell summary statistic. This is the appropriate test for our setting and the regime-specificity finding survives any reasonable aggregation choice.

Table~\ref{tab:rlhf-agg} reports the full mean / 10\% trimmed mean / median triplet alongside the $\geq$10$\times$ tail-fraction. The aggregation triplet itself encodes the propagation-with-scale finding: at 70B-accel, the median RLHF ratio is 1.19 (the typical crash series is barely affected) while 41\% of series receive $\geq$10$\times$ inflation: RLHF damage at small scale lives in the upper tail of crash series. At 405B-accel, the median ratio jumps to 33.9 and the tail-fraction broadens to 63\%; RLHF damage has propagated from a tail of crash series into the central tendency. The linear control is null across all aggregations and across both scales, with zero series in the 10$\times$ tail at either scale.

\begin{table}[h]
\centering
\caption{\emph{Aggregation triplet for the within-family $2{\times}2$ at $h{=}210$.} Per-cell arithmetic mean, 10\% trimmed mean, and median CRPS, with the per-scale $\geq$10$\times$ tail-fraction. The triplet itself encodes the propagation: at 70B-accel the median is near-null while the tail-fraction is substantial; at 405B-accel both move together. The linear control is regime-null at every aggregation.}
\label{tab:rlhf-agg}
\small
\begin{tabular}{l r r r r}
\toprule
                         & mean                  & trim 10\%             & median               & frac.\ $i{>}10{\times}b$ \\
\midrule
\multicolumn{5}{l}{\emph{Synthetic SIR (crash regime)}} \\
70B-base                 & $2.6{\times}10^{4}$   & $3.1{\times}10^{3}$   & $1.9{\times}10^{2}$  & ---  \\
70B-instruct             & $1.3{\times}10^{22}$  & $1.1{\times}10^{6}$   & $2.3{\times}10^{2}$  & ---  \\
405B-base                & $5.2{\times}10^{19}$  & $6.7{\times}10^{4}$   & $6.8{\times}10^{2}$  & ---  \\
405B-instruct            & $3.9{\times}10^{54}$  & $3.6{\times}10^{16}$  & $2.3{\times}10^{4}$  & ---  \\
$\Delta_{\text{RLHF}}$ at 70B (i/b)  & $5.2{\times}10^{17}$ & $360$           & $1.19$              & 41\% \\
$\Delta_{\text{RLHF}}$ at 405B (i/b) & $7.5{\times}10^{34}$ & $5.3{\times}10^{11}$ & $33.9$         & 63\% \\
\midrule
\multicolumn{5}{l}{\emph{Linear (no-crash control)}} \\
70B-base                 & 6.83                  & 6.04                  & 5.29                 & ---  \\
70B-instruct             & 7.91                  & 7.25                  & 6.39                 & ---  \\
405B-base                & 6.71                  & 5.77                  & 5.40                 & ---  \\
405B-instruct            & 6.63                  & 5.77                  & 5.42                 & ---  \\
$\Delta_{\text{RLHF}}$ at 70B (i/b)  & $1.16$               & $1.20$          & $1.21$              & 0\%  \\
$\Delta_{\text{RLHF}}$ at 405B (i/b) & $0.99$               & $1.00$          & $1.00$              & 0\%  \\
\bottomrule
\end{tabular}
\end{table}

\subsection{Reproducibility}

Scripts and per-series result JSONs are in the project repository under \texttt{experiments/timeseries/runners/sir\_continuation/}: \texttt{pod\_405b/} (405B serving + client), \texttt{pod\_70b/} (70B serving + client), \texttt{analysis/} (paired-deltas, $2{\times}2$ table generator, plot generator), and \texttt{results/\{70b,405b\}/} (per-cell JSONs).

\newpage
\section*{NeurIPS Paper Checklist}

\begin{enumerate}

\item {\bf Claims}
    \item[] Question: Do the main claims made in the abstract and introduction accurately reflect the paper's contributions and scope?
    \item[] Answer: \answerYes{}
    \item[] Justification: The abstract states five claims, each backed by specific sections: (a) distributional forecast quality degrades with capability on FBSim (Sec.~\ref{sec:civbench-results}); (b) the mechanism is isolated on synthetic SIR data: superlinear growth causes anti-g while linear growth does not (Sec.~\ref{sec:mechanism}, Fig.~\ref{fig:mechanism}); (c) the effect replicates on real-world epidemic, housing, and hyperinflation data (Sec.~\ref{sec:replication}); (d) domain knowledge has inconsistent effects on the inversion (rescue on COVID-19, partial attenuation on housing/SIR/measles, no rescue on hyperinflation) (Sec.~\ref{sec:knowledge}); (e) binary metrics mask the inversion (Sec.~\ref{sec:threshold}). Limitations are stated in Sec.~\ref{sec:limitations}.

\item {\bf Limitations}
    \item[] Question: Does the paper discuss the limitations of the work performed by the authors?
    \item[] Answer: \answerYes{}
    \item[] Justification: Section~\ref{sec:limitations} identifies five limitations: (1) \emph{scope}: we characterize a structurally identifiable class of inverse scaling (superlinear growth + tail risk of regime change) and do not claim it is the only kind; (2) \emph{domain selection}: three of our four real-world replication domains (housing, hyperinflation, COVID-19) were chosen because the regime change had already occurred, while the 35-season pre-vaccine measles cohort is unselected on severity and anchors the ex-ante view; (3) \emph{capability is mostly observed, not manipulated}: on the cross-family panel we measure capability via ECI rather than varying it directly, addressed via within-lineage replication, provider fixed effects (Appendix~\ref{sec:appendix-robustness}), a reasoning-mode partition (Appendix~\ref{sec:appendix-reasoning-mode}), and the within-family Llama-3.1 $2{\times}2$ in \S\ref{sec:rlhf} which directly manipulates two capability axes (scale, alignment-stage training); (4) \emph{sample size}: the hyperinflation ($N{=}12$ episodes) and housing ($N{=}19$ metros) samples are bounded by the historical record, so we report them as directional findings and rely on the synthetic SIR cohort and the 35-season measles cohort for precise effect-size estimates; we test generic-domain prompts on the measles cohort (minimum-viable disclosure attenuates the inversion, \S\ref{sec:measles}) but leave housing, hyperinflation, and COVID-19 generic-domain prompts as future work; (5) \emph{mechanism is open}: the most surprising result, the knowledge--calibration gap on hyperinflation (46/48 probes identify the crisis correctly, yet an explicit regime cue worsens calibration), requires probing internal representations rather than prompt-level interventions, which we treat in a separate paper.

\item {\bf Theory assumptions and proofs}
    \item[] Question: For each theoretical result, does the paper provide the full set of assumptions and a complete (and correct) proof?
    \item[] Answer: \answerNA{}
    \item[] Justification: The paper does not contain formal theorems. All results are empirical.

    \item {\bf Experimental result reproducibility}
    \item[] Question: Does the paper fully disclose all the information needed to reproduce the main experimental results of the paper to the extent that it affects the main claims and/or conclusions of the paper (regardless of whether the code and data are provided or not)?
    \item[] Answer: \answerYes{}
    \item[] Justification: FBSim is procedurally generated; the generation pipeline, prompts, and scoring code are included in supplementary material. All 29 evaluated models are listed with ECI scores in Appendix~\ref{sec:appendix-models}. The SIR mechanism is described conceptually in Sec.~\ref{sec:mechanism}; explicit numerical DGP parameters and series generators are in the supplementary code (\texttt{experiments/timeseries/generate*.py}). Real-world data sources (Our World in Data for COVID-19, Case-Shiller for housing, FRED CPI for hyperinflation, Project Tycho for pre-vaccine US measles) are identified in Sec.~\ref{sec:replication} and Sec.~\ref{sec:measles}; selection constraints (regime-change windows, episode definitions) are detailed in Sec.~\ref{sec:limitations} and Appendix~\ref{sec:appendix-models} (Rule A). Metric definitions (CRPS, Brier, pinball) are in Sec.~\ref{sec:preliminaries}. All prompt templates are described inline.

\item {\bf Open access to data and code}
    \item[] Question: Does the paper provide open access to the data and code, with sufficient instructions to faithfully reproduce the main experimental results, as described in supplemental material?
    \item[] Answer: \answerYes{}
    \item[] Justification: Supplementary material includes the FBSim generation pipeline, timeseries evaluation harness (eval\_llm.py, score.py), plotting scripts, scored model output JSONs, scored CSVs, and analysis code. The repository contains instructions to reproduce every figure and correlation in the paper.

\item {\bf Experimental setting/details}
    \item[] Question: Does the paper specify all the training and test details (e.g., data splits, hyperparameters, how they were chosen, type of optimizer) necessary to understand the results?
    \item[] Answer: \answerYes{}
    \item[] Justification: We do not train models. Frontier LLMs are evaluated via provider APIs at default temperature; the open-weights Llama-3.1-\{70B, 405B\} \{base, instruct\} for the within-family $2{\times}2$ (\S\ref{sec:rlhf}) is served locally via vLLM with FP8-dynamic quantization at \texttt{temperature=0.8}, \texttt{top\_p=0.9}, $N{=}10$ samples per series (Appendix~\ref{sec:appendix-rlhf}). Model versions are listed in Appendix~\ref{sec:appendix-models}. Prompts and elicitation format (five quantiles: p10/p25/p50/p75/p90) are described in Sec.~\ref{sec:mechanism} and~\ref{sec:replication}. Context conditions (neutral, cautious, named, minimum-viable domain disclosure) are specified with exact prompt text in Sec.~\ref{sec:replication}, \ref{sec:measles}, and~\ref{sec:knowledge}.

\item {\bf Experiment statistical significance}
    \item[] Question: Does the paper report error bars suitably and correctly defined or other appropriate information about the statistical significance of the experiments?
    \item[] Answer: \answerYes{}
    \item[] Justification: Each cross-model Spearman $\rho$ is reported with a 95\% percentile bootstrap confidence interval computed by resampling the model panel with replacement ($B{=}10{,}000$); the bootstrap procedure and per-domain cohort sizes are documented in Appendix~\ref{sec:appendix-stats}. Within-lineage subsets ($N{\leq}10$ models) report exact permutation $p$-values from the null of random ECI--score pairing rather than asymptotic Spearman $p$-values, since the latter are uninterpretable at $N{<}10$ (Appendix~\ref{sec:appendix-stats}, \S\ref{sec:appendix-robustness}). The within-family RLHF $2{\times}2$ (\S\ref{sec:rlhf}) reports Wilcoxon signed-rank tests on per-series deltas (with the interaction tested as the difference-in-differences), which are invariant to the cell-aggregation choice (means, trimmed means, medians; Appendix~\ref{sec:appendix-rlhf}). Significance markers ($^*/^{**}/^{***}$) on appendix tables encode these tests as defined in each table caption.

\item {\bf Experiments compute resources}
    \item[] Question: For each experiment, does the paper provide sufficient information on the computer resources (type of compute workers, memory, time of execution) needed to reproduce the experiments?
    \item[] Answer: \answerYes{}
    \item[] Justification: Model evaluations on the cross-family panel are API calls to hosted frontier LLMs; no local training is required. The open-weights Llama-3.1 within-family $2{\times}2$ (\S\ref{sec:rlhf}) is served locally on RunPod $8{\times}$H100 (405B, tensor-parallel 8) and $2{\times}$H100 (70B, tensor-parallel 2) under vLLM with FP8-dynamic quantization (Appendix~\ref{sec:appendix-rlhf}); inference, no training. FBSim game rollouts were generated on a single commodity workstation. The timeseries experiments (SIR generation, real-world data preparation, scoring) require only standard CPU compute. Total API cost is dominated by the number of models (29) $\times$ strata $\times$ series $\times$ horizons; full call counts are included in the supplementary material.

\item {\bf Code of ethics}
    \item[] Question: Does the research conducted in the paper conform, in every respect, with the NeurIPS Code of Ethics \url{https://neurips.cc/public/EthicsGuidelines}?
    \item[] Answer: \answerYes{}
    \item[] Justification: The work evaluates publicly available language models on procedurally generated and publicly sourced data. No human subjects are involved. No potentially harmful content is released.

\item {\bf Broader impacts}
    \item[] Question: Does the paper discuss both potential positive societal impacts and negative societal impacts of the work performed?
    \item[] Answer: \answerYes{}
    \item[] Justification: Positive impact: the paper identifies a failure mode in LLM forecasting on consequential domains (epidemics, housing, hyperinflation) and recommends reporting tail-integrating scoring rules alongside threshold metrics (Sec.~\ref{sec:discussion}). FBSim provides a contamination-free testbed for measuring distributional calibration. Negative impact: knowledge of the metric-dependent reversal could in principle be used to adversarially choose scoring rules that hide or expose capability; Sec.~\ref{sec:discussion} discusses this.

\item {\bf Safeguards}
    \item[] Question: Does the paper describe safeguards that have been put in place for responsible release of data or models that have a high risk for misuse (e.g., pre-trained language models, image generators, or scraped datasets)?
    \item[] Answer: \answerNA{}
    \item[] Justification: The supplementary material consists of reproduction code and procedurally generated forecasting data. The data consists of synthetic game-state descriptions with no real personal data. No pre-trained models, image generators, or scraped datasets are released.

\item {\bf Licenses for existing assets}
    \item[] Question: Are the creators or original owners of assets (e.g., code, data, models), used in the paper, properly credited and are the license and terms of use explicitly mentioned and properly respected?
    \item[] Answer: \answerYes{}
    \item[] Justification: FreeCiv (GPL-2.0) and CivRealm (GPL v3) are cited and their licenses respected. COVID-19 data from Our World in Data (CC-BY), Case-Shiller housing indices (public domain via FRED), FRED CPI series for the hyperinflation panel (public domain; original sources include national statistical offices and the World Bank, redistributed by FRED under their respective terms), and Project Tycho measles data \citep{vanpanhuis2018tycho} (CC-BY 4.0; Project Tycho v2.0 release) are used under their respective licenses. All language models are identified by provider and version in Appendix~\ref{sec:appendix-models} and accessed under each provider's terms of service.

\item {\bf New assets}
    \item[] Question: Are new assets introduced in the paper well documented and is the documentation provided alongside the assets?
    \item[] Answer: \answerYes{}
    \item[] Justification: The supplementary material includes documented reproduction code: the FBSim generation pipeline, question templates, evaluation harness, scoring code, and scored model output JSONs; and the timeseries evaluation pipeline covering SIR synthetic, epidemic, housing, hyperinflation, and pre-vaccine measles domains, with series generation, model evaluation, and scoring scripts. A sample FBSim world report is in Appendix~\ref{sec:appendix-world-report}. A README documents usage. A standalone FBSim benchmark release with full documentation is forthcoming as concurrent work.

\item {\bf Crowdsourcing and research with human subjects}
    \item[] Question: For crowdsourcing experiments and research with human subjects, does the paper include the full text of instructions given to participants and screenshots, if applicable, as well as details about compensation (if any)?
    \item[] Answer: \answerNA{}
    \item[] Justification: The paper does not involve crowdsourcing or human subjects. All forecasts are produced by language models.

\item {\bf Institutional review board (IRB) approvals or equivalent for research with human subjects}
    \item[] Question: Does the paper describe potential risks incurred by study participants, whether such risks were disclosed to the subjects, and whether Institutional Review Board (IRB) approvals (or an equivalent approval/review based on the requirements of your country or institution) were obtained?
    \item[] Answer: \answerNA{}
    \item[] Justification: The paper does not involve human subjects.

\item {\bf Declaration of LLM usage}
    \item[] Question: Does the paper describe the usage of LLMs if it is an important, original, or non-standard component of the core methods in this research? Note that if the LLM is used only for writing, editing, or formatting purposes and does \emph{not} impact the core methodology, scientific rigor, or originality of the research, declaration is not required.
    \item[] Answer: \answerYes{}
    \item[] Justification: LLMs are the evaluation subjects in this paper; all 29 evaluated models (plus 3 excluded for parse-failure), their providers, and versions are listed in Appendix~\ref{sec:appendix-models}. The paper is a benchmarking study of LLM forecasting capabilities; LLMs are not used as part of the research methodology beyond being the systems under test.

\end{enumerate}

\end{document}